%% file: CHEF-ICML2021.tex
\documentclass{article}
\usepackage[accepted]{icml2021}

\input{math_commands.tex}

\usepackage[utf8]{inputenc} % allow utf-8 input
\usepackage[T1]{fontenc}    % use 8-bit T1 fonts
\usepackage{hyperref}       % hyperlinks
\usepackage{url}            % simple URL typesetting
\usepackage{booktabs}       % professional-quality tables
\usepackage{amsfonts}       % blackboard math symbols
\usepackage{nicefrac}       % compact symbols for 1/2, etc.
\usepackage{microtype}      % microtypography

\usepackage{float}
\usepackage{graphicx}

\usepackage{color}
\usepackage{amsmath}
\usepackage{amssymb}
\usepackage{amsthm}
\usepackage{mathtools}
\usepackage[mathscr]{eucal}% \mathscr{ABCDEFGHIJKLMNOPQRSTUVWXYZ}
\usepackage{bm}% bold math
\usepackage{bbm}
\usepackage{relsize}
\usepackage{graphics}
\usepackage{wrapfig}
\usepackage{multirow}
\usepackage{enumitem}
\usepackage{tablefootnote}

\usepackage{array}

\usepackage{pgf}
\usepackage{xinttools}%,pgfkeys,pgfmath}
\usepackage{tikz}
\usetikzlibrary{arrows.meta}
\usepackage{textcomp}
\usetikzlibrary{calc,shapes,arrows,positioning,automata,trees}
\usepackage{placeins} % Provides \FloatBarrier
\usepackage{tabularx}
\usepackage{graphicx}
\usepackage{subcaption} % subfigures
\usepackage{listings}
\usepackage{xargs,lipsum,caption,changepage,ifthen}
\usepackage[toc,page]{appendix}

\newtheorem*{theorem*}{Theorem}
\newtheorem*{definition*}{Definition}

\newcommand\Bv{\bm{v}}

\newcommand\Bx{\bm{x}}
\newcommand\By{\bm{y}}
\newcommand\Bz{\bm{z}}
\newcommand\BI{\bm{I}}

\newcommand\BU{\bm{U}}
\newcommand\BV{\bm{V}}
\newcommand\BW{\bm{W}}
\newcommand\BX{\bm{X}}
\newcommand\BY{\bm{Y}}
\newcommand\BZ{\bm{Z}}
\newcommand\BZe{\bm{0}}

 \newcommand{\dR}{\mathbb{R}}

 \newcommand{\cL}{\mathcal{L}}

\usepackage{pdflscape} % For landscape tables

\newcolumntype{R}[1]{>{\raggedright\arraybackslash}p{#1}}
\newcolumntype{C}[1]{>{\centering\arraybackslash}p{#1}}
\newcolumntype{L}[1]{>{\raggedleft\arraybackslash}p{#1}}

\usepackage{bigdelim}
\usepackage{colortbl} % colored cells
\definecolor{mColor1}{rgb}{0.95,0.95,0.95}

\usepackage{microtype}
\usepackage{graphicx}
\usepackage{booktabs} % for professional tables

\usepackage{hyperref}

\icmltitlerunning{Cross-Domain Few-Shot Learning by Representation Fusion}

\begin{document}

\twocolumn[
\icmltitle{Cross-Domain Few-Shot Learning by Representation Fusion}

% It is OKAY to include author information, even for blind
% submissions: the style file will automatically remove it for you
% unless you've provided the [accepted] option to the icml2021
% package.

% List of affiliations: The first argument should be a (short)
% identifier you will use later to specify author affiliations
% Academic affiliations should list Department, University, City, Region, Country
% Industry affiliations should list Company, City, Region, Country

% You can specify symbols, otherwise they are numbered in order.
% Ideally, you should not use this facility. Affiliations will be numbered
% in order of appearance and this is the preferred way.
\icmlsetsymbol{equal}{*}

\begin{icmlauthorlist}
\icmlauthor{Thomas Adler}{equal,jku}
\icmlauthor{Johannes Brandstetter}{equal,jku}
\icmlauthor{Michael Widrich}{jku}
\icmlauthor{Andreas Mayr}{jku}
\icmlauthor{David Kreil}{iarai}
\icmlauthor{Michael Kopp}{iarai}
\icmlauthor{G{\"u}nter Klambauer}{jku}
\icmlauthor{Sepp Hochreiter}{jku,iarai}
\end{icmlauthorlist}

\icmlaffiliation{jku}{ELLIS Unit Linz and LIT AI Lab, Institute for Machine Learning, Johannes Kepler University Linz, Austria}
\icmlaffiliation{iarai}{Institute of Advanced Research in Artificial Intelligence (IARAI)}

\icmlcorrespondingauthor{Thomas Adler}{adler@ml.jku.at}
\icmlcorrespondingauthor{Johannes Brandstetter}{brandstetter@ml.jku.at}

% You may provide any keywords that you
% find helpful for describing your paper; these are used to populate
% the "keywords" metadata in the PDF but will not be shown in the document
\icmlkeywords{Machine Learning, ICML}

\vskip 0.3in
]

% this must go after the closing bracket ] following \twocolumn[ ...

% This command actually creates the footnote in the first column
% listing the affiliations and the copyright notice.
% The command takes one argument, which is text to display at the start of the footnote.
% The \icmlEqualContribution command is standard text for equal contribution.
% Remove it (just {}) if you do not need this facility.

%\printAffiliationsAndNotice{}  % leave blank if no need to mention equal contribution
\printAffiliationsAndNotice{\icmlEqualContribution} % otherwise use the standard text.

\begin{abstract}
In order to quickly adapt to new data,
few-shot learning aims at learning from few examples, often by
using already acquired knowledge.
The new data often differs from the previously seen data
due to a domain shift, that is,
a change of the input-target distribution.
%The reason for observing new data is a domain shift, that is,
%a change of the input-target distribution.
While several methods perform well on small domain shifts like
new target classes with similar inputs,
larger domain shifts are still challenging.
Large domain shifts 
may result in high-level concepts 
that are not shared between the original and the new domain,
whereas low-level concepts like edges in images might 
still be shared and useful. % in the new domain.
For cross-domain few-shot learning,
we suggest representation fusion to 
unify different abstraction levels 
of a deep neural network into one representation.
We propose Cross-domain Hebbian Ensemble Few-shot learning (CHEF),
which achieves representation fusion by 
an ensemble of Hebbian learners 
acting on different layers 
of a deep neural network.
Ablation studies show that representation fusion 
is a decisive factor to boost cross-domain few-shot learning.
On the few-shot datasets
\textit{mini}Imagenet and \textit{tiered}Imagenet
with small domain shifts, CHEF is competitive
with state-of-the-art methods.
On cross-domain few-shot benchmark challenges with larger domain shifts,
CHEF establishes novel state-of-the-art results
in all categories.
We further apply CHEF on a real-world 
cross-domain application in drug discovery. 
We consider a domain shift from 
bioactive molecules to environmental 
chemicals and drugs with twelve associated toxicity prediction tasks.
%First a small domain shift is
%considered by classifying the biological 
%effect of compounds on a new target (bioassay) 
%with few measurements available.
%Then a large domain shift is considered,
%where compounds of particular chemical properties have
%to be classified although they were not present in training.
On these tasks, that are highly relevant for computational drug discovery,
CHEF significantly outperforms all its competitors.
\end{abstract}

\section{Introduction}

Currently, deep learning is criticized because it is
data hungry, has limited capacity for transfer,
insufficiently integrates prior knowledge, and
presumes a largely stable world \citep{Marcus:18}.
In particular, these problems appear after a domain shift,
that is, a change of the input-target distribution.
A domain shift forces deep learning models to adapt.
%After a domain shifts, the goal is to exploit models 
The goal is to exploit models
that were trained on the typically rich original data
for solving tasks from the new domain with much less data.
Examples for domain shifts are
new users or customers, new products and product lines,
%new document types,
new diseases (e.g.\ adapting from SARS to COVID19),
new images from another field (e.g.\ from cats to dogs % or from cats to birds 
or from cats to bicycles),
new social behaviors after societal change (e.g.\ introduction of cell phones, pandemic),
self-driving cars in new cities or countries (e.g.\ from European countries 
to Arabic countries), and robot manipulation of new objects.

Domain shifts are often tackled by 
meta-learning \citep{schmidhuber87, bengio90, hochreiter01}, 
since it exploits already acquired knowledge to adapt to new data.
One prominent application of meta-learning 
dealing with domain shifts is few-shot learning, %\citep{Hospedales:20},
%Meta-learning is often used for few-shot learning \citep{Hospedales:20},
%which is also important for domain shifts, 
since, typically, from
the new domain much less data is available than from the original domain.
%\citep{bendavid2010}.
Meta-learning methods perform well on small domain shifts like
new target classes with similar inputs.
However, larger domain shifts are still challenging for current approaches.
Large domain shifts lead to inputs, which are considerably different 
from the original inputs and possess different high-level concepts.
Nonetheless, low-level concepts are often still shared between the inputs of
the original domain and the inputs of the new domain.
For images, such shared low-level concepts
can be edges, textures, small shapes, etc.
One way of obtaining low level concepts
is to train a new deep learning model from scratch,
where the new data is merged with the original data.
%Low level concepts can for example be obtained by training
%a new deep learning model from scratch, where the new data
%is merged with the original data.
However, although models of the original domain are 
often available, the original data, 
which the models were trained on, often are not. 
%The original data may not be available for several reasons,
This might have several reasons,
e.g.\ the data owner does no longer grant access to the data, 
General Data Protection Regulation (GDPR)
does no longer allow access to the data,
IP restrictions prevent access to the data,
sensitive data items must not be touched anymore (e.g.\ phase III drug candidates),
or data is difficult to extract again.
We therefore suggest to effectively exploit original data models directly
by accessing not only high level but also low level abstractions.
In this context, we propose a cross-domain few-shot learning method
extracting information from different levels of 
abstraction in a deep neural network.

\textbf{Representation fusion.} 
Deep Learning constructs neural network models 
that represent the data at multiple levels of abstraction \citep{lecun2015deep}.
We introduce \textit{representation fusion},
which is the concept of unifying and merging information from different levels of abstraction.
Representation fusion uses 
a fast and adaptive system for detecting relevant information
at different abstraction levels of a deep neural network, 
which we will show allows solving versatile and complex cross-domain tasks.

\textbf{CHEF.} We propose cross-domain ensemble
few-shot learning (CHEF) that achieves representation fusion
by an ensemble of Hebbian learners, which are built upon a trained network.
CHEF naturally addresses the problem of domain shifts
which occur in a wide range of real-world applications.
Furthermore, since CHEF only builds on 
representation fusion,
it can adapt to new characteristics
of tasks like 
unbalanced data sets, 
classes with few examples,
change of the measurement method, 
new measurements in unseen ranges,
new kind of labeling errors,
and more. 
The usage of simple Hebbian learners 
allows the application of CHEF without needing to backpropagate information through the backbone network.

The main contributions of this paper are:
\begin{itemize}
    \item We introduce representation fusion as the concept of unifying and merging information from different layers of abstraction. An extensive ablation study shows that representation fusion is a decisive factor to boost cross-domain few-shot learning.
    \item We introduce CHEF\footnote{Our implementation is available at \href{https://github.com/ml-jku/chef}{github.com/ml-jku/chef}.} as our new cross-domain few-shot learning method that builds on representation fusion.
    We show that using different layers of abstraction allows one to successfully tackle various few-shot learning tasks across a wide range of different domains. CHEF does not need to backpropagate information through the backbone network.
    \item We apply CHEF to various cross-domain few-shot tasks and obtain several state-of-the-art results. We further apply CHEF to cross-domain real-world applications from drug discovery, where we outperform all competitors. 
\end{itemize}

\textbf{Related work. }
Representation fusion builds on learning a meaningful
representation \citep{bengio2013representation, girshick2014rich} at multiple levels of abstraction \citep{lecun2015deep, schmidhuber2015deep}.
The concept of using representations from different
layers of abstraction has been used in CNN architectures~\citep{lecun1998gradient}
such as \citet{huang2017densely, rumetshofer2018human, hofmarcher2019accurate}, 
in CNNs for semantic segmentation
in the form of multi-scale context pooling \citep{yu2015multi,chen2018encoder}, 
and in the form of context capturing
and symmetric upsampling \citep{ronneberger2015u}.
Learning representations from different domains has been explored 
by \citet{federici2020learning, tschannen2020on} under the viewpoint 
of mutual information optimization. 
Work on domain shifts discusses the problem that new inputs are considerably 
different from the original inputs \citep{kouw2019, kouw2018, webb2018, gama2014, widmer1996}. 
Domain adaptation \citep{pan2009, bendavid2010} overcomes this problem 
by e.g.\ reweighting the original samples \citep{huang2007}, 
learning features that are invariant to a 
domain shift \citep{ganin2016jmlr, xu2019cvpr}
or learning a classifier in the new domain. 
Domain adaptation where only few data
is available in the new domain \citep{bendavid2010, lu20} 
is called cross-domain few-shot learning \citep{guo19, lu20, tseng2020cross},
which is an instance of the general few-shot learning setting \citep{fei2006one}.
Few-shot learning can be roughly divided into three approaches \citep{lu20, Hospedales:20}:
(i) augmentation, (ii) metric learning, and (iii) meta-learning.
For (i), where the idea is to learn an augmentation to produce more than the few samples available, supervised \citep{dixit2017aga,kwitt2016one} and 
unsupervised \citep{hariharan2017low, pahde2019low, gao2018low} methods are considered. For (ii), approaches aim to learn a pairwise similarity metric under which similar samples obtain high similarity scores \citep{koch2015siamese, ye2018deep, hertz2006learning}.
For (iii), methods comprise \textit{embedding and nearest-neighbor} approaches
\citep{snell2017prototypical, sung18, vinyals16}, \textit{finetuning} approaches
\citep{finn17, rajeswaran19, ravi17, andrychowicz2016learning}, and \textit{parametrized} approaches \citep{gidaris18,ye20,lee19,yoon19,mishra18,hou19,rusu18}.
Few-shot classification under domain shifts for metric-based methods has been discussed in \citet{tseng2020cross}.
Ensemble methods for few-shot learning have been applied in \citet{dvornik2019diversity},
where an ensemble of distance-based classifiers is designed
from different networks. In contrast, our method builds an ensemble of different layers from the same network. Hebbian learning as part of a few-shot learning method has been implemented in \citet{munkhdalai2018metalearning},
where fast weights that are used for binding labels to representations
are generated by a Hebbian learning rule.

\section{Cross-domain few-shot learning} \label{sec:cross-domain}

\textbf{Domain shifts.}
We assume to have data $(\Bx,\By)$, where $\Bx \in \BX$ is the input data
and $\By \in \BY$ is the target data.
A domain is a distribution $p$ over $\BX\times \BY$
assigning each pair $(\Bx,\By)$ a probability $p(\Bx,\By)$.
A domain shift is a change from $p(\Bx,\By)$ to $\tilde{p}(\Bx,\By)$.
We measure the magnitude of the domain shift by
a distance $d(p,\tilde{p})$ between the distributions $p$ and $\tilde{p}$.
We consider four types of domain shifts \citep{kouw2019, kouw2018, webb2018, gama2014, widmer1996}:
\begin{itemize}
    \item \textbf{Prior shift (small domain shift)}: $p(\By)$ is changed to $\tilde{p}(\By)$, while $p(\Bx \mid \By)$ stays the same. For example, when new classes are considered (typical case in few-shot learning):
        $p(\Bx,\By) = p(\By)p(\Bx \mid \By)$ and $\tilde{p}(\Bx,\By) = \tilde{p}(\By)p(\Bx \mid \By)$.
    \item \textbf{Covariate shift (large domain shift)}: $p(\Bx)$ is changed to $\tilde{p}(\Bx)$, while $p(\By \mid \Bx)$ stays the same. For example, when new inputs are considered, which occurs when going from color to grayscale images, using a new measurement device, or looking at traffic data from different continents: 
        $p(\Bx,\By) = p(\Bx)p(\By \mid \Bx)$ and $\tilde{p}(\Bx,\By) = \tilde{p}(\Bx)p(\By \mid \Bx)$.
    \item \textbf{Concept shift}:  $p(\By \mid \Bx)$ is changed to $\tilde{p}(\By \mid \Bx)$, while $p(\Bx)$ stays the same. For example, when including new aspects changes the decision boundaries:
        $p(\Bx,\By) = p(\Bx)p(\By \mid \Bx)$ and $\tilde{p}(\Bx,\By) = p(\Bx)\tilde{p}(\By \mid \Bx)$.
    \item \textbf{General domain shift}: domain shift between $p(\Bx,\By)$ and $\tilde{p}(\Bx,\By)$. For example, going from Imagenet data to grayscale X-ray images (typical case in cross-domain datasets).
\end{itemize}

\textbf{Domain shift for images.}
We consider the special case that the input $\Bx$ is an image.
In general, domain shifts can be measured on the raw image distributions e.g.\ 
by using the $\mathcal{H}$-divergence \citep{bendavid2010}. 
However, distances between raw image distributions were shown to be 
less meaningful in computer vision tasks than abstract representations 
of deep neural networks \citep{heusel2017, salismans2016}. We approximate the 
distance between the joint distributions $d(p(\Bx,\By),\tilde{p}(\Bx,\By))$ 
by the distance between the marginals $d(p(\Bx),\tilde{p}(\Bx))$, 
which is exact in the case of the covariate shift for certain choices of $d(\cdot,\cdot)$, 
like e.g.\ the Jensen-Shannon divergence.
To measure the distance between the marginals $d(p(\Bx),\tilde{p}(\Bx))$ we use the 
Fr\'{e}chet Inception Distance (FID; \citealp{heusel2017}), i.e.\ the Wasserstein-2
distance of the features of the respective images activating an Inception v3 network 
\citep{szegedy2016rethinking} under a Gaussian assumption. The FID has proven 
reliable for measuring performance of Generative Adversarial Networks~\citep{goodfellow2014generative}.

\textbf{Cross-domain few-shot learning.}
Large domain shifts lead to inputs, which are considerably 
different from the original inputs.
As a result, the model trained on the original domain
will not work anymore on the new domain.
To overcome this problem, domain adaptation 
techniques are applied \citep{pan2009, bendavid2010}.
Domain adaption can be achieved
in several ways, e.g.\ by reweighting the original
samples \citep{huang2007}.
Another possibility is to learn a classifier in the new domain.
Domain adaptation where in the new domain 
only few data is available \citep{bendavid2010} which can be used for learning
is called cross-domain few-shot learning \citep{guo19, lu20, tseng2020cross}.
In an $N$-shot $K$-way few-shot learning setting, 
the training set (in meta learning also called one episode)
consists of $N$ samples for each of the $K$ classes.

\section{Cross-domain Hebbian Ensemble Few-shot learning (CHEF)} \label{sec:chef}

\begin{figure*}[t]
\centering
\includegraphics[width=.77\linewidth]{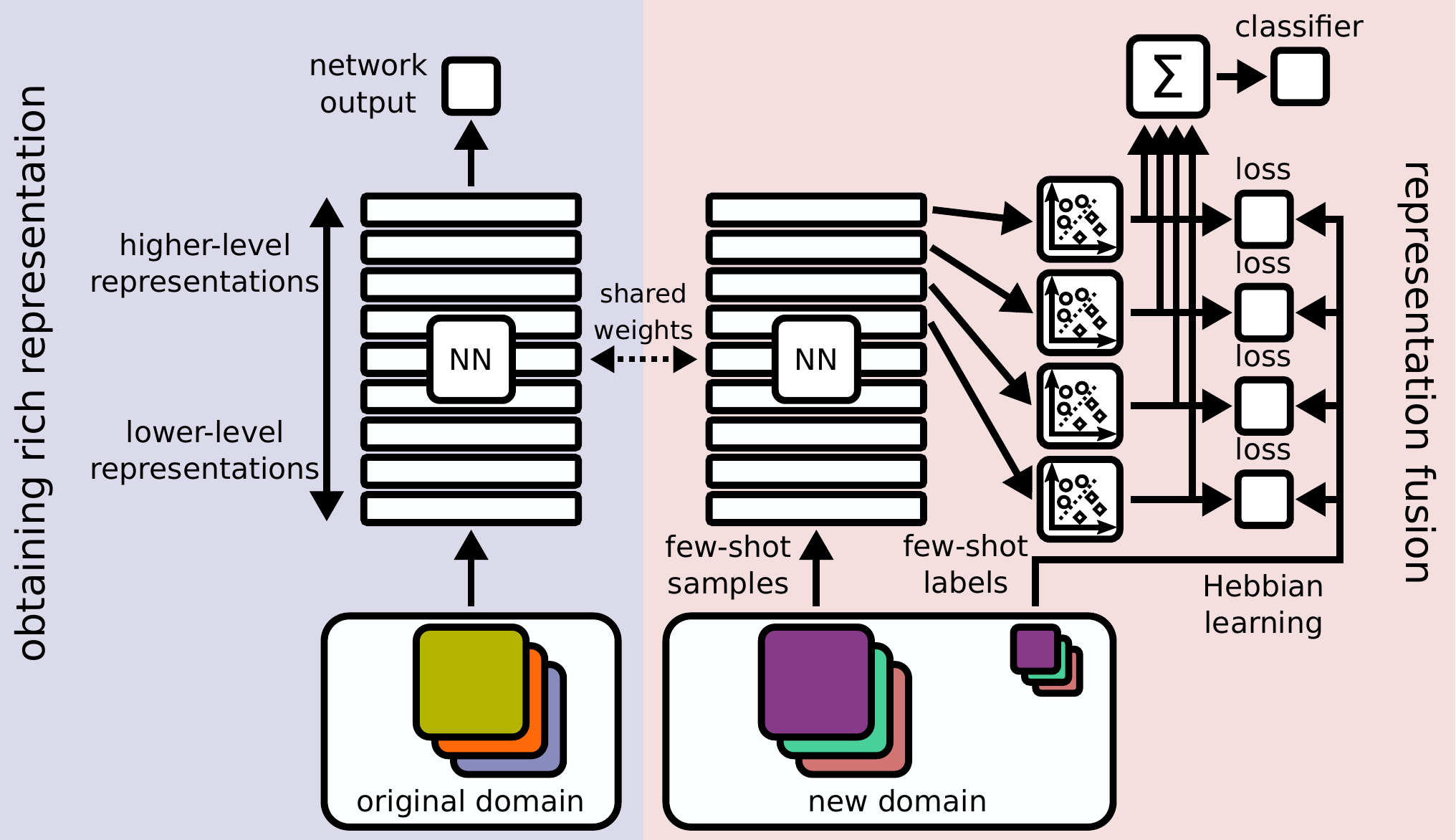}
\caption[]{Working principle of CHEF. 
An ensemble of Hebbian learners is applied to the upper layers 
of a trained neural network. Distilling information from 
different layers of abstraction is called representation fusion.
Each Hebbian learner is iteratively
optimized and the results are combined. 
CHEF does not require backpropagation of error signals through the entire
backbone network, only the parameters of the individual learners need adjustment.
Hebbian learners can easily be exchanged with other supervised techniques.}
\label{fig:sketch_system1system}
\end{figure*}

We propose a new cross-domain few-shot learning method, CHEF, 
that consists of an ensemble of Hebbian learners built on representation fusion (see Fig.~\ref{fig:sketch_system1system}). 
%Figure~\ref{fig:sketch_system1system} sketches our
%CHEF approach.
A schematic view of the CHEF algorithm is outlined in Alg.~\ref{alg:chef}.

We conducted an extensive ablation study on the cross-domain 
few-shot learning challenge datasets~\citep{guo19} to test (i) the influence of 
different levels of abstraction 
(Tab.~\ref{tab:ablation2} in the appendix)
and (ii) the influence of different learning algorithms 
(Tab.~\ref{tab:cross_domain_res18} in the appendix).
%For (i), the general insight is that
%the larger the domain shift, the higher the influence of low level features becomes.
%Combining the last six layers via representation fusion always achieves the best results.
For (ii), we compared support vector machines (SVMs) with different kernels, 
random forest (RF) with different numbers of trees, 
$k$-nearest neighbors with different numbers of neighbors, 
and a Hebbian learning rule 
(Tab.~\ref{tab:cross_domain_res18} in the appendix).
In principle, any learning algorithm can be used for representation fusion.
We chose Hebbian learning since it is a simple and fast learning rule
that outperformed or performed on par with the other methods 
(Tab.~\ref{tab:cross_domain_res18} in the appendix).

%To obtain predictions for the individual levels of abstraction we could 
%in principle use any learning algorithm. We choose a Hebbian 
%learning rule because it is simple and fast while being robust and reliable. 
%However, the choice of this algorithm can change with the problem setting. 
%An ensemble of Hebbian learners is applied to the upper layers 
%of a trained neural network. Each Hebbian learner is iteratively
%optimized and the results are combined.

\textbf{Representation fusion. }
%CHEF is build on the concept of representation fusion.
%Deep learning \citep{lecun2015deep} models provide hierarchical 
%representations that allow to fuse information from different
%layers of abstraction. There exists 
%a wide variety of methods to obtain such general and rich representations. 
%These range from plain supervised learning over multi-task 
%learning~\citep{caruana97} to unsupervised methods like data imputation~\citep{rubin2004multiple} or auto encoders \citep{ballard1987modular}.
CHEF builds its ensemble of learners
using representation fusion.
Deep learning models \citep{lecun2015deep} provide hierarchical 
representations that allow to fuse information from different
layers of abstraction.
The closer a layer is to the network output the more specific are its features. 
Conversely, the closer a layer is to the input of the network, 
the more general are the features. 
%Therefore, combining information from different layers means to combine 
%features of different specificity to the task at hand. 
In cross-domain few-shot learning, 
it is not a priori clear how specific or general the features should be 
because this depends on how close the target domain is to the training domain. Therefore, we design our few-shot learning algorithm 
such that it can flexibly choose the specificity of the features depending on 
the current episode. 
We achieve this by representation fusion,
where the learners are  applied to 
several layers at different levels of the backbone network in parallel.
This yields a separate prediction for each level of abstraction. 
The final classification result is then obtained from the sum of logits 
arising from the respective learners.
Ablation studies revealed that combining the last six layers via representation fusion achieves the best results.

\textbf{Computation time. }
In contrast to many other methods, CHEF does not require
backpropagation of error signals through the entire backbone network.
Only the parameters of the learners that are obtained for the 
uppermost layers need adjustment. This makes CHEF extremely fast and versatile.
A detailed study of the computation time can be found in the appendix.

\textbf{Obtaining one Hebbian Learner. } \label{hebbian_learning}
We consider an $N$-shot $K$-way few-shot learning setting. 
Let $\Bz_i \in \dR^D$ be a feature 
vector obtained from activating a pre-trained backbone network with a sample 
$\Bx_i$ up to a certain layer, where $D$ is the number of 
units in that layer. 
We combine the $NK$ feature 
vectors into a matrix $\BZ \in \dR^{NK \times D}$ and 
initialize a weight matrix 
$\BW \in \dR^{K \times D}$. 
The Hebbian learning rule~\citep{hebb1949} reads:
\begin{equation}
    \Delta w_{ij} = \alpha \cdot a_i \cdot a_j \ ,
\end{equation}
where $\Delta w_{ij}$ is the weight change from neuron $i$ to neuron $j$,
$a_i$ and $a_j$ are the activations of neuron $i$ and neuron $j$,
respectively, and $\alpha$ is the Hebbian learning rate.
In accordance with \citet{fregnac2002hebbian}, 
we use the update rule
\begin{equation}
\BW \gets \BW - \alpha \BV^\top \BZ \label{eqn:hebb}
\end{equation}
for a given number of steps, where
$\BV \in \dR^{NK \times K}$
is the matrix of postsynaptic responses $\Bv_i$. 
We design the Hebbian learning rate and the number of steps for which 
to run the update rule \ref{eqn:hebb} as hyperparameters of our method. 
Given a loss function $\cL(\cdot, \cdot)$ and few-shot labels 
$\By_i$, we choose the postsynaptic response
\begin{equation}
\Bv_i = \nabla_{\hat{\By}} \cL(\By_i, \hat{\By})|_{\hat{\By} = \BW \Bz_i},
\end{equation}
casting the update rule \ref{eqn:hebb} effectively as a gradient descent step. 
However, since we use only a few steps of rather strong updates we prefer to 
view this as a Hebbian learning rule. 
We initialize the weight matrix $\BW$ with zeros. 
In principle, any other initialization is possible but due to strong updates 
the initialization scheme is of minor importance. 

\begin{algorithm}
\caption{CHEF algorithm. The data matrix $\BX$ consists of 
input vectors $\Bx_i$ and the label matrix $\BY$ consists of the corresponding 
label vectors $\By_i$. 
The function BB activates the backbone network up
to a certain layer specified by an index $l$. 
$L$ is the set of indices specifying 
the layers used in the ensemble,
$\cL$ is the loss function of the few-shot learning task at hand.
The function \textsc{HebbRule} executes $M$ steps of the Hebbian 
learning rule and yields a weight matrix $\BW$ 
that maps the feature vectors in $\BZ$ to vectors of length $K$, 
which are then used for $K$-fold classification. 
$\alpha$ is the Hebbian learning rate.}
\begin{algorithmic}
\REQUIRE $\alpha, M, \operatorname{Softmax}(\cdot), \cL(\cdot, \cdot), \operatorname{BB}(\cdot,\cdot)$
\FUNCTION{$\operatorname{HebbRule}(\BX, \BY, l)$}%{$\BX, \BY, l$}
  \STATE $\BW \gets \BZe, \BZ \gets \operatorname{BB}(\BX, l)$
  \FOR{$m \in \{1, \dots, M\}$} 
    \STATE $\BV \gets \nabla_{(\BZ \BW^\top)} \cL(\BY, \operatorname{Softmax}(\BZ \BW^\top))$
    \STATE $\BW \gets \BW - \alpha \BV^\top \BZ$
  \ENDFOR
  \STATE {\bfseries return} $\BW$
\ENDFUNCTION
\FUNCTION{$\operatorname{Ensemble}(\BX, \BY, L)$}%{$\BX, \BY, L$}
  \STATE $\BU = \sum_{l \in L} \operatorname{BB}(\BX, l)\,\operatorname{HebbRule}(\BX, \BY, l)^\top$
  \STATE {\bfseries return} $\operatorname{Softmax}(\BU)$
\ENDFUNCTION
\end{algorithmic}
\label{alg:chef}
\end{algorithm}

\section{Experiments}\label{sec:experiments}

We apply CHEF to four cross-domain few-shot challenges,
where we obtain state-of-the-art results in all categories.
The four cross-domain few-shot challenges are 
characterized by domain shifts of different magnitude,
which we measure using the Fr\'{e}chet-Inception-Distance (FID).
We conduct ablation studies showing the influence of the 
different layer representations on the results.
Further, we test CHEF on two standardized image-based few-shot classification 
benchmark datasets established in the field,
which are characterized by a prior domain shift: \textit{mini}Imagenet \citep{vinyals16} 
and \textit{tiered}Imagenet \citep{ren18}.
Finally, we illustrate the impact of our CHEF approach on two real-world 
applications in the field of drug discovery,
which are characterized first by a small domain shift and 
second by a large domain shift.

\subsection{Cross-domain few-shot learning} \label{sec:cross_domain}

\textbf{Dataset and evaluation. }
The cross-domain few-shot learning challenge \citep{guo19} uses 
\textit{mini}Imagenet as training 
domain and then evaluates the trained models on four different test domains
with increasing distance to the training domain: 
1)~CropDisease \citep{mohanty2016} consisting of 
plant disease images, 2)~EuroSAT \citep{helber2019eurosat}, 
a collection of satellite images, 
3)~ISIC2018 \citep{tschandl2018ham10000, codella2019skin} containing
dermoscopic images of skin lesions, and 
4)~ChestX \citep{wang2017chestx} containing a set of X-ray images. 
%%%%%%%%%%%%%555
%For evaluation protocols, we follow~\citet{rusu18}
%and measure the accuracy by
%For evaluation, we measure the accuracy by
%drawing 800 $N$-shot $K$-way tasks from the respective cross-domain test set.
For evaluation, we measure the accuracy
drawing 800 tasks (five test instances per class) 
from the cross-domain test set.
%Each task consists of a support set $\cD_t^s$ 
%of $N$ labeled images drawn from each of the $K$ randomly chosen 
%classes and a query set $\cD_t^q$ 
%of 15 query images per class. 
Following prior
work, we focus on 5-way/5-shot, 5-way/20-shot, and 5-way/50-shot tasks.
We report the average accuracy and a 95\,\% confidence 
interval across all test images and tasks.

Recent work \citep{ericsson2020selfsupervised,cai2020sbmtl,wang2020comparison} 
showed impressive results in the meta-transfer learning setting,
where pre-trained models are optimized over specific few-shot learning tasks
via transfer learning. In contrast, our experiments focus on the case where only one 
episode of the 5-way/\{5,20,50\}-shot data of the new domain is available for 
adjusting the model. 
That is, we relinquish episodic training completely, which is in 
accordance with the findings of \citet{laenen2020episodes}. 

\textbf{Measuring the domain shifts via FID. } 
In \citet{guo19}, the four datasets of the new domain are characterized 
by their distance to the original domain using three criteria:
whether images contain perspective distortion, 
the semantic content of images, 
and color depth.
In Table~\ref{tab:fid_result}, we provide measurements of
the domain shift of these four datasets 
with respect to the original \textit{mini}Imagenet dataset
using the FID. 
The FID measurements confirm the characterization in \citet{guo19},
except that the EuroSAT dataset is closer to the original domain
than the CropDisease dataset. The difference in both FID measurements
is mostly driven by the mean terms.
This can be explained by the fact that the FID does not measure perspective distortion 
and satellite images might have a higher variety of shapes and colors than plant images.
Recent associative techniques, e.g.~\citep{ramsauer2020hopfield} are not tested so far,
but are a potentially promising filed of future research.

\begin{table*}[t]
    \centering
    \begin{tabular}{llc}
         \toprule
         Dataset & Conceptual difference to original domain (\textit{mini}Imagenet) & FID \\ 
         \midrule
         CropDisease & None & 257.58  \\
         EuroSAT & No perspective distortion & 151.64  \\
         ISIC2018 & No perspective distortion, unnatural content & 294.05  \\
         ChestX & No perspective distortion, unnatural content, different color depth & 312.52  \\
         \bottomrule
    \end{tabular}
    \caption{Conceptual difference and domain shift between \textit{mini}Imagenet and the four cross-domain datasets CropDisease, EuroSAT, ISIC2018, and ChestX. The domain shift is measured using the FID.}
    \label{tab:fid_result}
    %\vspace{-1cm}
\end{table*}

\textbf{CHEF implementation. }
We perform pre-training on the \textit{mini}Imagenet dataset
similar but not identical to that in \citet{ye20}. 
We utilize a softmax output layer with as many units as classes 
are contained in the meta-training and the meta-validation sets combined. 
We make a validation split on the combination of these two sets for 
supervised learning, i.e.\ instead of separating whole classes into 
the validation set (vertical split) we move a randomly selected fraction of samples 
of each class into the validation set (horizontal split) as it is standard in 
supervised learning. We evaluate CHEF using the same ResNet-10 backbone architecture 
as in \citet{guo19}.
For better representation fusion, we place two fully 
connected layers after the last convolutional layer.
We perform model selection during training using 
the cross-entropy loss function on the horizontal data split,
and perform hyperparameter selection for CHEF on the vertical data split.

\begin{figure}[!htp]
 \centering
 \includegraphics[width=0.88\linewidth]{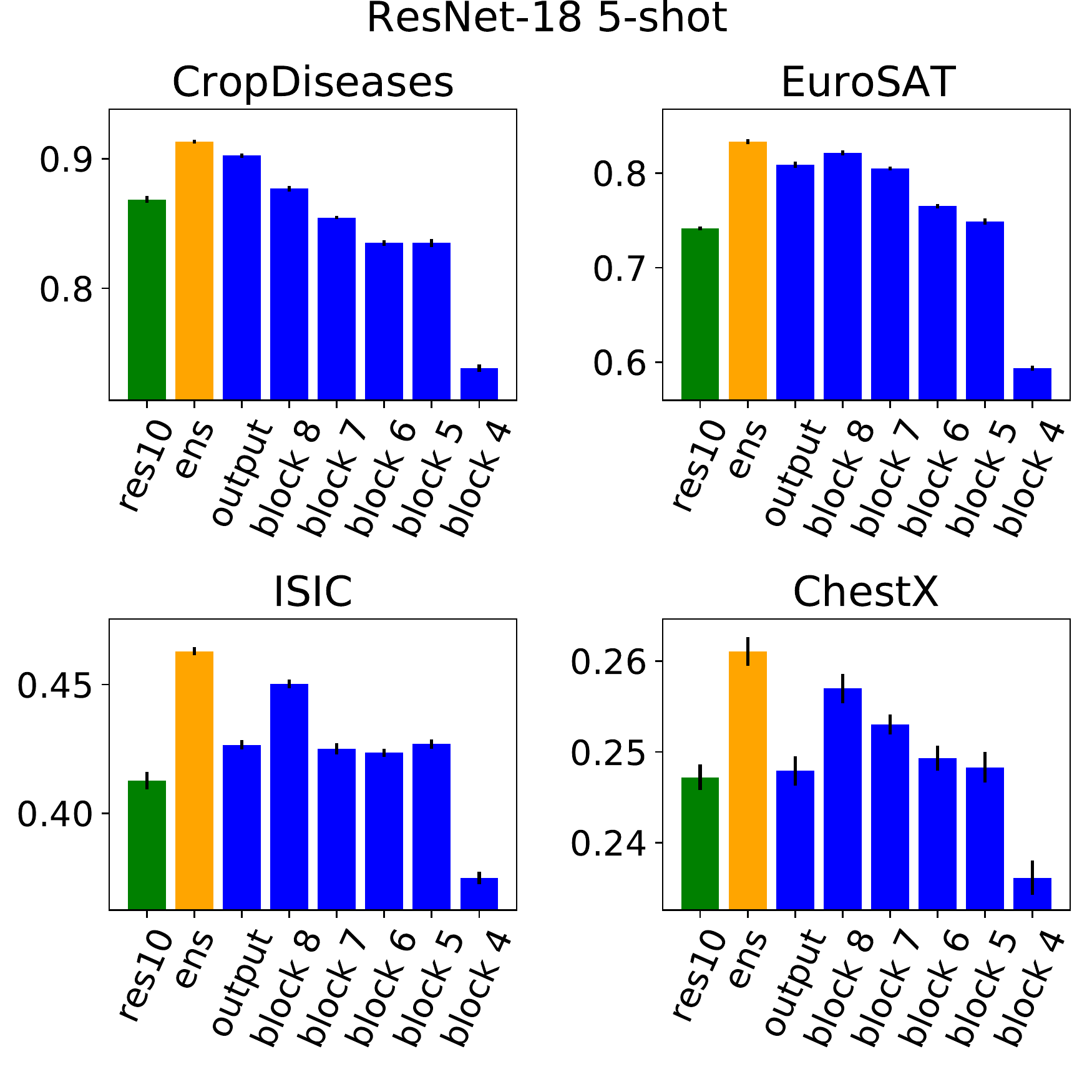}
 \includegraphics[width=0.88\linewidth]{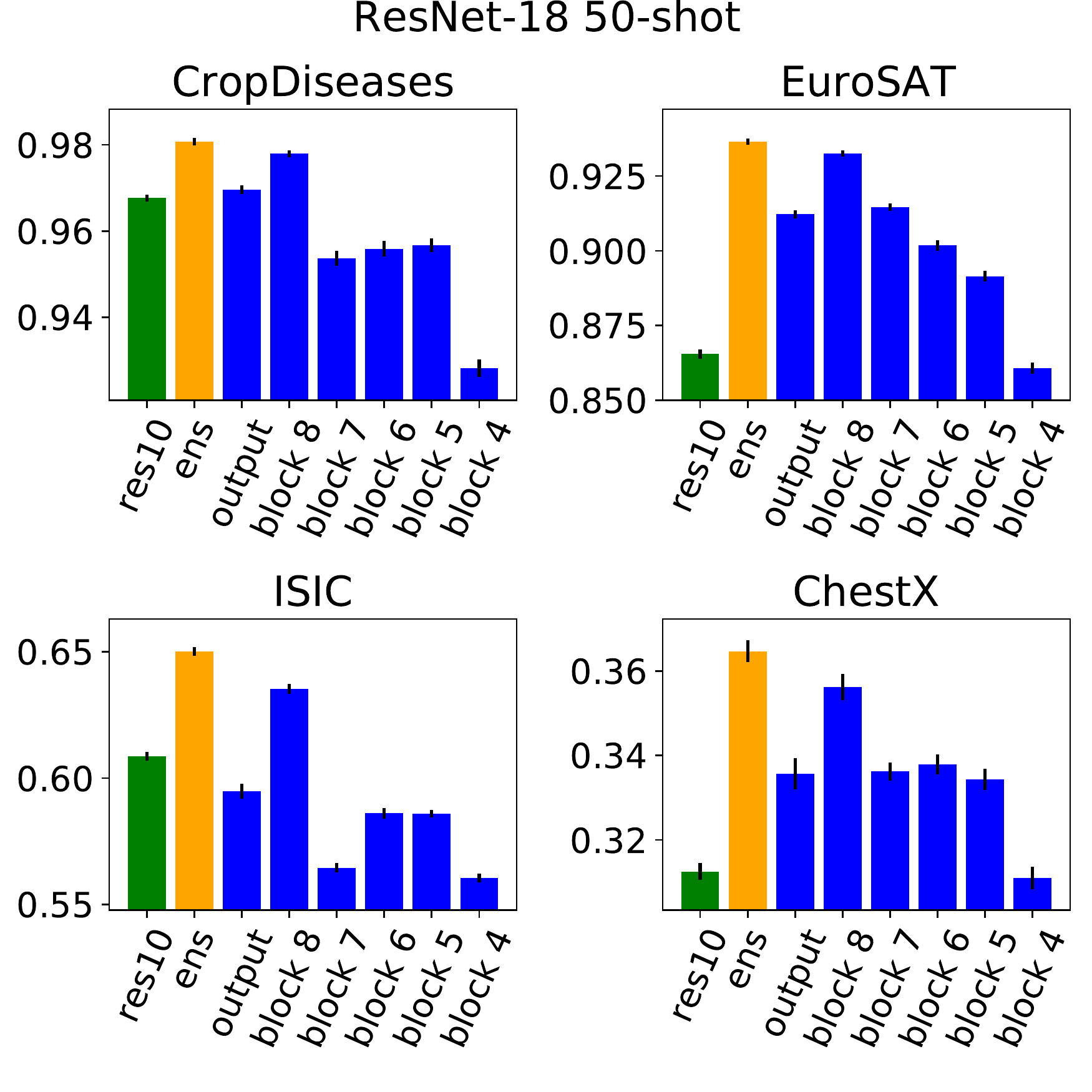}
 \caption{Ablation study of 5-shot and 50-shot top-1 accuracies 
     (along with $95\%$ confidence intervals) of different 
     residual blocks and the output layer of 
     a PyTorch Imagenet-pretrained ResNet-18 and the ensemble result (orange, ``ens'') 
     on the four different datasets of the cross-domain few-shot learning 
     benchmark. For comparison, also the ResNet-10 ensemble results (green) 
     are included. The ResNet-18 is pretrained on the whole Imagenet dataset.}
 \label{fig:res18}
\end{figure}

\textbf{Results and ablation study. }CHEF achieves state-of-the-art performance in all 12 categories. 
Results are provided in Table~\ref{tab:meta}.
An ablation of the influence of representation fusion
on 5-way 5-shot and 50-shot results
is shown in Fig.~\ref{fig:res18} where we use a pre-trained PyTorch \citep{paszke2019:pytorch} ResNet-18, which consists of 8 convolutional blocks. 5-way 20-shot results can be found in Fig.~\ref{fig:res18_20shot} in the appendix.
Results are obtained by applying our Hebbian learning 
rule to the logits of the output layer, to the pre-activations of the 
blocks 4 through 8 individually, and to an ensemble of them.
The results are considerably better than the above reported ResNet-10 results, which presumably arises from the fact that the power of representation fusion is larger since the ResNet-18 is pretrained on the whole Imagenet dataset,
and, thus, has stronger representations in the individual layers.
This illustrates the power of CHEF considering better feature abstractions. Another interesting insight is that for the ChestX dataset, the dataset with the largest domain shift, the lower level features gain importance.
In contrast, in CropDiseases and EuroSAT, the datasets with the smallest domain shifts, the differences between block 4 and block 8 are large. 
In general, the farther the domain is away from the original domain the more 
important are features from lower layers, i.e.\ features that are less specific 
to the original domain. Since CHEF combines features of different specificity to the training domain, it is particularly powerful in cross-domain settings.

\begin{table*}[!htb]
\centering
\scalebox{0.82}{
\begin{tabular}{l c c c c c c}
    \hline
    %\hline
&\multicolumn{3}{c}{\textbf{CropDiseases 5-way}}& \multicolumn{3}{c}{\textbf{EuroSAT 5-way}}  \\
    %\cline{2-7}
    \textbf{Method} & \textbf{5-shot} &\textbf{20-shot}&\textbf{50-shot}&\textbf{5-shot} &\textbf{20-shot} &\textbf{50-shot} \\
    \hline%\cline{2-7}
MatchingNet$^{\dagger}$   &66.39 $\pm$ 0.78  & 76.38 $\pm$ 0.67& 58.53 $\pm$ 0.73& 64.45 $\pm$ 0.63&77.10 $\pm$ 0.57&54.44 $\pm$ 0.67 \\

MatchingNet+FWT$^{\dagger}$    & 62.74 $\pm$ 0.90 & 74.90 $\pm$ 0.71 & 75.68 $\pm$ 0.78& 56.04 $\pm$ 0.65&  63.38 $\pm$ 0.69 & 62.75 $\pm$ 0.76 \\

MAML$^{\dagger}$   & 78.05 $\pm$ 0.68 & \underline{89.75} $\pm$ 0.42 &-& 71.70 $\pm$ 0.72&81.95 $\pm$ 0.55& -\\
  
ProtoNet$^{\dagger}$    & \underline{79.72} $\pm$ 0.67 &88.15 $\pm$ 0.51 &90.81 $\pm$ 0.43 &\underline{73.29} $\pm$ 0.71 & \underline{82.27} $\pm$ 0.57 & 80.48 $\pm$ 0.57 \\

ProtoNet+FWT$^{\dagger}$    & 72.72 $\pm$ 0.70 &85.82 $\pm$ 0.51&87.17 $\pm$ 0.50& 67.34 $\pm$ 0.76& 75.74 $\pm$ 0.70  & 78.64 $\pm$ 0.57\\

RelationNet$^{\dagger}$    &  68.99 $\pm$ 0.75 &80.45 $\pm$ 0.64&85.08 $\pm$ 0.53& 61.31 $\pm$ 0.72  & 74.43 $\pm$ 0.66& 74.91 $\pm$ 0.58 \\

RelationNet+FWT$^{\dagger}$    & 64.91 $\pm$ 0.79  & 78.43 $\pm$ 0.59& 81.14 $\pm$ 0.56& 61.16 $\pm$ 0.70 & 69.40 $\pm$ 0.64 & 73.84 $\pm$ 0.60 \\

MetaOpt$^{\dagger}$   & 68.41 $\pm$ 0.73&82.89 $\pm$ 0.54 & \underline{91.76} $\pm$ 0.38& 64.44 $\pm$ 0.73 &79.19 $\pm$ 0.62 & \underline{83.62} $\pm$ 0.58  \\

CHEF (Ours)  &\textbf{86.87} $\pm$ 0.27 &\textbf{94.78} $\pm$ 0.12&\textbf{96.77} $\pm$ 0.08 &\textbf{74.15} $\pm$ 0.27 & \textbf{83.31} $\pm$ 0.14 & \textbf{86.55} $\pm$ 0.15\\

%\emph{Ours (Res18)}  &91.33 $\pm$ 0.0 &96.87 $\pm$ 0.0&97.84 $\pm$ 0.0 &83.48 $\pm$ 0.0 &90.91 $\pm$ 0.0& 93.04 $\pm$ 0.0\\

%    \hline
%\end{tabular}}
%\end{table*}

%\begin{table*}[!t!]
%\centering
%\scalebox{0.66}{
%\begin{tabular}{c c c c | c c c}
    \hline
    \hline
    \hline
 &\multicolumn{3}{c}{\textbf{ISIC 5-way}}& \multicolumn{3}{c}{\textbf{ChestX 5-way}}  \\
    %\cline{2-7}
    \textbf{Method} & \textbf{5-shot} &\textbf{20-shot}&\textbf{50-shot}&\textbf{5-shot} &\textbf{20-shot} &\textbf{50-shot} \\
    \hline%\cline{2-7}
MatchingNet$^{\dagger}$    & 36.74 $\pm$ 0.53& 45.72 $\pm$ 0.53&  54.58 $\pm$ 0.65& 22.40 $\pm$ 0.7 & 23.61 $\pm$ 0.86 &22.12 $\pm$ 0.88 \\

MatchingNet+FWT$^{\dagger}$    &  30.40 $\pm$ 0.48 & 32.01 $\pm$ 0.48 & 33.17 $\pm$ 0.43 & 21.26 $\pm$ 0.31& 23.23 $\pm$ 0.37  & 23.01 $\pm$ 0.34 \\

MAML$^{\dagger}$   &\underline{40.13} $\pm$ 0.58 &\underline{52.36} $\pm$ 0.57 & - & 23.48 $\pm$ 0.96 & 27.53 $\pm$ 0.43  & - \\
  
ProtoNet$^{\dagger}$    & 39.57 $\pm$ 0.57&49.50 $\pm$ 0.55&51.99 $\pm$ 0.52& \underline{24.05} $\pm$ 1.01& \underline{28.21} $\pm$ 1.15 &29.32 $\pm$ 1.12 \\

ProtoNet+FWT$^{\dagger}$    &  38.87 $\pm$ 0.52 & 43.78 $\pm$ 0.47 & 49.84 $\pm$ 0.51& 23.77 $\pm$ 0.42 & 26.87 $\pm$ 0.43 & \underline{30.12} $\pm$ 0.46 \\

RelationNet$^{\dagger}$    & 39.41 $\pm$ 0.58 & 41.77 $\pm$ 0.49 &49.32 $\pm$ 0.51 & 22.96 $\pm$ 0.88&26.63 $\pm$ 0.92&28.45 $\pm$ 1.20 \\

RelationNet+FWT$^{\dagger}$    &35.54 $\pm$ 0.55 & 43.31 $\pm$ 0.51& 46.38 $\pm$ 0.53&22.74 $\pm$ 0.40 & 26.75 $\pm$ 0.41& 27.56 $\pm$ 0.40 \\

MetaOpt$^{\dagger}$   &36.28 $\pm$ 0.50 &49.42 $\pm$ 0.60& \underline{54.80} $\pm$ 0.54 & 22.53 $\pm$ 0.91 &25.53 $\pm$ 1.02& 29.35 $\pm$ 0.99\\

CHEF (Ours)  &\textbf{41.26} $\pm$ 0.34 &\textbf{54.30} $\pm$ 0.34 & \textbf{60.86} $\pm$ 0.18 &\textbf{24.72} $\pm$ 0.14 & \textbf{29.71} $\pm$ 0.27& \textbf{31.25} $\pm$ 0.20\\

%\emph{Ours (Res18)}  &45.12 $\pm$ 0.0 &56.89 $\pm$ 0.0&63.11 $\pm$ 0.0 &25.85 $\pm$ 0.0 &31.87 $\pm$ 0.0& 36.50 $\pm$ 0.0\\
    \hline
    \multicolumn{4}{l}{\footnotesize{$^\dagger$ Results reported in \citet{guo19}}} \\
\end{tabular}}
\caption{Comparative results of few-shot learning methods on four proposed cross-domain few-shot challenges CropDiseases, EuroSAT, ISIC, and ChestX. The average 5-way few-shot classification accuracies ($\%$, top-1) along with $95\%$ confidence intervals are reported on the test split of each dataset. Runner-up method is underlined.}
\label{tab:meta}
\end{table*}

\subsection{\textit{mini}Imagenet and \textit{tiered}Imagenet}

\textbf{Datasets and evaluation.}
The \textit{mini}Imagenet dataset \citep{vinyals16}
consists of 100 randomly chosen classes from the ILSVRC-2012 dataset~\citep{russakovsky15}.
%with random splits 
%into 64, 16 and 20 classes
%for meta-training, meta-validation, and meta-testing, respectively.
%Each class contains 600 images of size 84×84. 
We use the commonly-used class split proposed in~\citet{ravi17}. 
The \textit{tiered}Imagenet dataset \citep{ren18} is a subset
of ILSVRC-2012 \citep{russakovsky15}, composed of 608 classes grouped in 34 high-level categories. 
%It is composed of 608 classes grouped
%into 34 high-level categories,
%which are divided into 20, 6, and 8 categories 
%for meta-training, meta-validation, and meta-testing, respectively.
%This corresponds to 351, 97 and 160 classes 
%for meta-training, meta-validation, and
%meta-testing, respectively, all with images of size 84 × 84.
%The \textit{tiered}Imagenet dataset aims to minimize the
%semantic similarity between the splits. \\
%%%%%%%%%%%%%%%%%%%%%%%%%
%For evaluation protocols, we follow~\citet{rusu18}
%and measure the accuracy by
For evaluation, we measure the accuracy
drawing 800 tasks (five test instances per class) 
from the meta-test set.
%Each task consists of a support set $\cD_t^s$ 
%of $N$ labeled images drawn from each of the $K$ randomly chosen 
%classes and a query set $\cD_t^q$ 
%of 15 query images per class. 
Following prior
work, we focus on 5-way/1-shot and 5-way/5-shot tasks.
We report the average accuracy and a 95\,\% confidence 
interval across all test images and tasks.

\textbf{CHEF implementation and results.} 
We perform pre-training of the respective backbone networks
on the \textit{mini}Imagenet
and the \textit{tiered}Imagenet
dataset in the same way as described in 
Sec.~\ref{sec:cross_domain}.
We evaluate CHEF 
using two different backbone architectures: 
a Conv-4 and a ResNet-12 network. We use the Conv-4 network described 
by \citet{vinyals16}.
%which is a stack of 4 modules, each of 
%which consists of a $3 \times 3$ convolutional layer with 64 units, a 
%batch normalization layer~\citep{ioffe15}, a ReLU activation and 
%$2 \times 2$ max-pooling layer. 
Following \citet{lee19}, we 
configure the ResNet-12 backbone as 4 residual blocks, 
%with 64, 160, 320, and 640 units, respectively. 
which contain a 
max-pooling and a batch-norm layer and are regularized by 
DropBlock \citep{ghiasi18}.
Again, model selection and hyper-parameter tuning is performed as 
described in Sec.~\ref{sec:cross_domain}.
%Again, we perform model selection during pre-training 
%and hyperparameter tuning after pre-training as 
%described in Sec.~\ref{sec:cross_domain}.
%
%In order to allow for meaningful representation fusion we put two linear layers on top of the backbone architecture.
%with block sizes of $1 \times 1$ for the 
%first two blocks and $5 \times 5$ for the latter two blocks, all at 
%a dropout rate of $0.1$. 
%For both backbone networks 
%we use the Adam~\citep{kingma15} optimizer 
%with a learning rate of $10^{-3}$. 
%We perform model selection during training using 
%the cross-entropy loss function on the horizontal data split.
%We perform hyperparameter selection for CHEF on the meta-validation set.
%
%\textbf{Results. }
CHEF achieves state-of-the-art performance in 5 out of 8 categories.
Results are provided in Table~\ref{tab:main}.
An ablation study of the \textit{mini}Imagenet and \textit{tiered}Imagenet results can be found in the appendix.

\begin{table*}[!htb]
\centering
\scalebox{0.78}{
\begin{tabular}{lccccc}
\hline 
 & & \multicolumn{2}{c}{\textbf{\textit{mini}Imagenet 5-way}} & \multicolumn{2}{c}{\textbf{\textit{tiered}Imagenet 5-way}}  \\
\textbf{Method} & \textbf{Backbone} & \textbf{1-shot} & \textbf{5-shot} & \textbf{1-shot} & \textbf{5-shot}  \\
\hline 
MatchingNet \citep{vinyals16} & Conv-4 & $43.56\pm0.84$ & $55.31\pm0.73$ & - & - \\
Meta-LSTM \citep{ravi17} & Conv-4 & $43.44\pm0.77$ & $60.60\pm0.71$ & - & - \\
MAML \citep{finn17} & Conv-4 & $48.70\pm1.84$ & $63.11\pm0.92$ & $51.67\pm1.81^{\dagger}$ & $70.30\pm1.75^{\dagger}$  \\
ProtoNets \citep{snell17} & Conv-4 & $49.42\pm0.78$ & $68.20\pm0.66$ & $48.58\pm0.87^{\dagger}$ & $69.57\pm0.75^{\dagger}$ \\
Reptile \citep{nichol18} & Conv-4 & $47.07\pm0.26$ & $62.74\pm0.37$ & $48.97\pm0.21^{\dagger}$ & $66.47\pm0.21^{\dagger}$ \\
RelationNet \citep{sung18} & Conv-4 & $50.44\pm0.82$ & $65.32\pm0.70$ & $54.48\pm0.93^{\dagger}$ & $71.32\pm0.78^{\dagger}$ \\
IMP \citep{allen19} & Conv-4 & $49.60\pm0.80$ & $68.10\pm0.80$ & - & - \\
%SIB \citep{hu20} & Conv-4 & $58.0\pm0.6$ & $70.7\pm0.4$ & - & - \\ TRANSDUCTIVE
FEAT \citep{ye20} & Conv-4 & $55.15\pm0.20$ & $71.61\pm0.16$ & - & - \\ %NEED ALL TRAINING DATA
Dynamic FS \citep{gidaris18} & Conv-4 & $56.20\pm0.86$ & $\underline{72.81}\pm0.62$ & - & - \\ %NEED ALL TRAINING DATA
MetaQCD \citep{metaQDA2021} & Conv-4 & ${\underline{56.32}\pm0.79}$ & ${72.54\pm0.23}$ & ${\underline{58.11}\pm0.58}$ & ${\underline{74.26}\pm0.69}$ \\
CHEF (Ours) & Conv-4 & $\mathbf{57.60}\pm0.29$ & $\mathbf{73.26}\pm0.23$ & $\mathbf{61.10}\pm0.21$ & $\mathbf{75.83}\pm0.25$ \\
\hline
\hline
\hline
SNAIL \citep{mishra18} & ResNet-12 & $55.71\pm0.99$ & $68.88\pm0.92$ & - & - \\
Dynamic FS \citep{gidaris18} & ResNet-12 & $55.45\pm0.89$ & $70.13\pm0.68$ & - & - \\
TADAM \citep{oreshkin18} & ResNet-12 & $58.50\pm0.30$ & $76.70\pm0.30$ & - & - \\
MTL \citep{sun19} & ResNet-12 & $61.20\pm1.80$ & $75.50\pm0.80$ & - & - \\
VariationalFSL \citep{zhang19} & ResNet-12 & $61.23\pm0.26$ & $77.69\pm0.17$ & - & - \\
TapNet \citep{yoon19} & ResNet-12 & $61.65\pm0.15$ & $76.36\pm0.10$ & $63.08\pm0.15$ & $80.26\pm0.12$ \\
%LEO \citep{rusu18} & ResNet-12 & $61.76\pm0.08$ & $77.59\pm0.12$ & $66.33\pm0.05$ & $81.44\pm0.09$ \\ USES WRN NETWORK
MetaOptNet \citep{lee19} & ResNet-12 & $62.64\pm0.61$ & $78.63\pm0.46$ & $65.81\pm0.74$ & $81.75\pm0.53$ \\
CTM \citep{li19} & ResNet-12 & $62.05\pm0.55$ & $78.63\pm0.06$ & $64.78\pm0.11$ & $81.05\pm0.52$ \\
%MetaOptNet \citep{lee19} & ResNet-12 & $64.09\pm0.62$ & $80.00\pm0.45$ & %$62.64\pm0.61$ & $78.63\pm0.46$ \\
CAN \citep{hou19} & ResNet-12 & $63.85\pm0.48$ & $79.44\pm0.34$ & $69.89\pm0.51$ & $84.23\pm0.37$ \\%NEED ALL TRAINING DATA
FEAT \citep{ye20} & ResNet-12 & $\mathbf{66.78}\pm0.20$ & $\mathbf{82.05}\pm0.14$ & $\mathbf{70.80}\pm0.23$ & $\underline{84.79}\pm0.16$ \\%NEED ALL TRAINING DATA\\
%NEED ALL TRAINING DATA
CHEF (Ours) & ResNet-12 & $\underline{64.11}\pm0.32$ & $\underline{79.99}\pm0.21$ & $\mathbf{70.70}\pm0.35$ & $\mathbf{85.97}\pm0.09$ \\
%Ours (Res18) & ResNet-18 & $91.24$ & $98.16$ & $86.56$ & $95.78$ \\
\hline
\multicolumn{4}{l}{\footnotesize{$^\dagger$ Results reported in~\citep{liu19}}} \\
%\multicolumn{5}{l}{\footnotesize{$^\ddagger$ Information is shared among test examples using batch normalization (transductive)~\citep{liu19}}} \\
%\multicolumn{4}{l}{\footnotesize{$^\sharp$ Learner uses all training data}}
\end{tabular}}
\caption{Comparative results of few-shot learning methods on the two benchmark datasets \textit{mini}Imagenet and \textit{tiered}Imagenet. The average 5-way few-shot classification accuracies ($\%$, top-1) along with $95\%$ confidence intervals are reported on the test split of each dataset. Runner-up method is underlined.}
\label{tab:main}
\vspace{-0.16in}
\end{table*}

\newpage
\subsection{Example Application: Drug Discovery}\label{sec:drugdiscovery}

In drug discovery, it is essential to know properties of drug candidates, 
such as biological activities %(\emph{bioactivities}) 
or toxicity \citep{sturm2020industry,mayr2018large}. 
Since the measurements of these properties
require time- and cost-intensive laboratory experiments, 
machine learning models are used to substitute such
measurements \citep{hochreiter2018machine}.
However, due to the high experimental effort 
often only few high-quality measurements are available
for training. 
Thus, few-shot learning is highly relevant 
for computational drug discovery.
We introduce a new perspective to the problem
by viewing the few high-quality measurements as a domain shift
of a much larger dataset.

\textbf{Problem setting. }
We consider a 50-shot cross-domain few-shot learning setting
in the field of toxicity prediction, utilizing
the \emph{Tox21 Data Challenge dataset} (Tox21) 
with twelve different toxic effects \citep{huang2017editorial, mayr2016deeptox}.
Around 50 available measurements is a typical scenario 
when introducing a new high-quality
assay in drug design. 
So far, the standard approach
to deal with such few data points
was to use machine learning methods like
Support Vector Machines (SVMs; \citealp{cortes1995support}) 
or Random Forests (RFs; \citealp{breiman2001random}).
However, these methods do not exploit
the rich data available, like the \emph{ChEMBL20 
drug discovery benchmark} (ChEMBL20) \citep{mayr2018large,gaulton2017chembl}.
Viewing the Tox21 data as a domain shift of the 
ChEMBL20 allows the application 
of cross-domain few-shot learning methods.
In this setting,
a domain shift can be observed
both in the input data and in the 
target data.
The molecules (input domain) are 
strongly shifted towards a specialized 
chemical space, with a Jaccard index of $0.01$
between the two datasets, and 
the biological effects (output domain) 
are shifted towards toxicity without any overlap in this domain. 
A further shift is in the distribution of 
the target labels, which are now much more imbalanced in comparison
to ChEMBL20.
In order to mirror this distribution shift correctly,
the number of toxic vs.\ non-toxic molecules 
in the training sets
for each of the twelve few-shot tasks (twelve different toxic effects) 
are sub-sampled accordingly.
For example, the 50-shot few-shot scenario
(50-50 toxic/non-toxic)
is adjusted to
a 10-90 scenario.  
For the twelve few-shot learning tasks, 
training samples are drawn from the 
training set and 
test samples are drawn from the
test set of the Tox21 data, respectively.
We sample individually
for each of the twelve tasks.

\textbf{CHEF implementation for molecules.}
We first train
a fully-connected deep neural network (FCN)
for the prediction of bioactivities from the 
ChEMBL20 database (original domain).
The network is trained in a massive multi-task setting,
where 1,830 
tasks are predicted at once, such that the network
is forced to learn proficient representations that can be shared for multiple tasks \citep{ma2015deep,unterthiner2014multi}.
The total number of 892,480 features of the ChEMBL20 database
was reduced by a sparseness criterion on the molecules
to 1,866 features.
The neurons in the input layer of the FCN 
represent one of 1,866 ECFP6 \citep{rogers2010extended} features, 
which are used as a feature representation for describing 
the raw structure of the molecules.
Each neuron of the output layer represents one of the
1,830 prediction tasks.
We use the pre-trained network and apply CHEF
by representation fusion of the 
three bottleneck layers of the network
for predicting the twelve different 
toxic effects of the new domain of the Tox21 Data Challenge.

%\newpage

%\begin{wraptable}[15]{r}{0.38\textwidth}
\begin{table}[t]
    \centering
    %\vspace{-0.25cm}
    \begin{tabular}{lc}
         \toprule
         Method & ROC-AUC \\ 
         \midrule
         CHEF & {\bf 0.76} $\pm$ 0.02   \\
         SVM & 0.66  $\pm$ 0.03 \\
         RF & 0.64$\pm$ 0.03 \\
         \bottomrule
    \end{tabular}
    \caption{
    ROC-AUC performance for few-shot drug discovery.
    CHEF is compared to conventional methods (SVM, RF)
    for the prediction of toxic effects.
    Mean and standard deviation are computed across twelve
    different effects and across 100 differently sampled
    training and test sets.}
    \label{tab:result_dd}
    %\vspace{1cm}
%\end{wraptable}
\end{table}

\textbf{Experimental evaluation.} 
We evaluate the performance of CHEF 
on the twelve tasks of the Tox21 Data Challenge
and compare it to conventional methods, like SVMs and RFs, 
that are used in drug design when few data is available.
We use SVMs with a MinMax kernel since it previously yielded
the best results \citep{mayr2018large}.
%which often works well, since
%it can take into account features, that are not observed
%in the training data.
For CHEF, only the 1,866 ECFP input features
of ChEMBL20 pre-training network database are used
where features with only few occurrences in the training set are discarded
since they do not give enough learning signal for neural network training.
For SVMs and RFs, all encountered ECFP features are used.
ROC-AUC values are computed across the twelve
tasks of the 
Tox21 Data Challenge and across 100
differently sampled training and test sets.
CHEF achieves significantly better
ROC-AUC values than SVMs and RFs.
Table~\ref{tab:result_dd} shows the results 
($p$-value~$< 10^{-17}$ for both SVM and RF 
when using a paired Wilcoxon test).
Results for the twelve individual tasks and
a more detailed description is given in the appendix.
CHEF significantly outperforms traditional methods in drug design,
which demonstrates the great potential of 
cross-domain few-shot learning in this field.

\section{Conclusion} 

We have introduced CHEF as new cross-domain few-shot learning method.
CHEF builds on the concept of representation fusion,
which unifies information from different levels of abstraction. 
An extensive ablation study shows that representation fusion is a decisive factor to boost cross-domain few-shot learning.
Representation fusion allows one to successfully tackle
various few-shot learning problems with large domain shifts across a wide range of different tasks
CHEF obtains new state-of-the-art results in all categories 
of the broader study of cross-domain few-shot learning benchmarks.
Finally, we have tested the performance of CHEF
in a real-world cross-domain application in drug discovery, 
i.e.\ toxicity prediction when a domain shift appears.
CHEF significantly outperforms all traditional approaches 
demonstrating great potential for applications in 
computational drug discovery.

\iffalse
\section*{Acknowledgments}
The ELLIS Unit Linz, the LIT AI Lab, the 
Institute for Machine Learning,
are supported by
the Federal State Upper Austria.
IARAI is supported by Here Technologies.
We thank the projects
AI-MOTION (LIT-2018-6-YOU-212),
DeepToxGen (LIT-2017-3-YOU-003),
AI-SNN (LIT-2018-6-YOU-214), 	
DeepFlood (LIT-2019-8-YOU-213),
Medical Cognitive Computing Center (MC3),
PRIMAL (FFG-873979),
S3AI (FFG-872172),
DL for granular flow (FFG-871302),
ELISE (H2020-ICT-2019-3 ID: 951847),
AIDD (MSCA-ITN-2020 ID: 956832).
We thank
Janssen Pharmaceutica,
UCB Biopharma SRL,
Merck Healthcare KGaA,
Audi.JKU Deep Learning Center,
TGW LOGISTICS GROUP GMBH,
Silicon Austria Labs (SAL),
FILL Gesellschaft mbH,
Anyline GmbH,
Google Brain,
ZF Friedrichshafen AG,
Robert Bosch GmbH,
T\"{U}V Austria,
and the NVIDIA Corporation.
\fi

\bibliography{icml2021}
\bibliographystyle{icml2021}

\newpage
\onecolumn
\appendix

\newpage
\section{Appendix}

\begin{table*}[!p]
\centering
\scalebox{1.}{
\begin{tabular}{l c c c c c c}
    \hline
&\multicolumn{3}{c}{\textbf{CropDiseases 5-way}}& \multicolumn{3}{c}{\textbf{EuroSAT 5-way}}  \\
    \textbf{Learner} & \textbf{5-shot} &\textbf{20-shot}&\textbf{50-shot}&\textbf{5-shot} &\textbf{20-shot} &\textbf{50-shot} \\
    \hline
Hebbian & 91.34 & 96.99 & 98.07 & 83.44 & 91.62 & 93.65 \\

SVM (linear) & 89.59 & 97.10 & 98.47 & 80.54 & 92.14 & 94.45 \\

SVM (RBF) & 67.08 & 96.36 & 97.72 & 57.82 & 89.95 & 92.82 \\

SVM (polynomial) & 80.61 & 96.04 & 97.63 & 60.99 & 90.63 & 93.37 \\

RF (10) & 78.40 & 92.81 & 95.53 & 70.29 & 84.27 & 88.35 \\

RF (50) & 89.27 & 96.45 & 97.51 & 80.07 & 89.39 & 91.73 \\

RF (100) & 90.65 & 96.71 & 97.65 & 81.60 & 90.23 & 92.12 \\

3-NN & 77.39 & 87.12 & 90.68 & 39.99 & 46.36 & 50.28 \\

5-NN & 75.28 & 86.56 & 90.27 & 37.27 & 44.03 & 48.22 \\

7-NN & 72.93 & 85.82 & 89.66 & 35.36 & 42.37 & 46.66 \\

    \hline
 &\multicolumn{3}{c}{\textbf{ISIC 5-way}}& \multicolumn{3}{c}{\textbf{ChestX 5-way}}  \\
    \textbf{Learner} & \textbf{5-shot} &\textbf{20-shot}&\textbf{50-shot}&\textbf{5-shot} &\textbf{20-shot} &\textbf{50-shot} \\
    \hline
Hebbian  & 46.29 & 58.85 & 65.01 & 26.11 & 31.83 & 36.47 \\

SVM (linear) & 30.85 & 59.27 & 67.46 & 18.39 & 29.33 & 37.83 \\

SVM (RBF) & 13.83 & 52.01 & 62.5 & 16.27 & 17.95 & 34.16 \\

SVM (polynomial) & 18.17 & 56.57 & 64.98 & 16.83 & 26.23 & 36.81 \\

RF (10) & 37.06 & 48.53 & 54.92 & 23.52 & 26.73 & 30.44 \\

RF (50) & 42.12 & 55.47 & 61.37 & 25.47 & 30.14 & 34.29 \\

RF (100) & 44.04 & 56.78 & 62.58 & 26.24 & 30.83 & 35.72 \\

3-NN & 35.24 & 45.68 & 51.16 & 24.47 & 26.89 & 28.99 \\

5-NN & 34.49 & 45.92 & 51.93 & 24.15 & 27.45 & 29.68 \\

7-NN & 32.86 & 45.92 & 51.99 & 24.57 & 26.98 & 30.32 \\

    \hline
    %\multicolumn{4}{l}{\footnotesize{$^\dagger$ Results reported in \citet{guo19}}} \\
\end{tabular}}
\caption{Results of our few-shot learning method CHEF on four proposed cross-domain few-shot challenges CropDiseases, EuroSAT, ISIC, and ChestX. We compare Hebbian learning to support vector machine (SVM) with linear, polynomial and RBF kernel, respectively alongside random forest (RF) with 10, 50, and 100 trees as well as $k$ nearest neighbor (NN) with $k=3$, $k=5$, and $k=7$, respectively. All learners were used on top of a ResNet-18 backbone network pre-trained on Imagenet. The average 5-way few-shot classification accuracies ($\%$, top-1) over 800 episodes are reported.}
\label{tab:cross_domain_res18}
\end{table*}

\begin{table*}[!p]
\centering
\scalebox{0.9}{
\begin{tabular}{l c c c c c c c}
    \hline
    \multicolumn{8}{c}{\textbf{CropDiseases 5-way/5-shot}} \\
    \textbf{Learner} & \textbf{Ensemble} & \textbf{Output} & \textbf{Block 8} & \textbf{Block 7} & \textbf{Block 6} & \textbf{Block 5} & \textbf{Block 4} \\
    \hline
Hebbian & 91.34 & 90.27 & 87.70 & 85.47 & 83.50 & 83.50 & 73.83 \\

SVM (linear) & 89.59 & 88.56 & 89.37 & 80.66 & 76.65 & 76.75 & 49.20 \\

SVM (RBF) & 67.08 & 69.00 & 57.85 & 34.54 & 25.94 & 26.21 & 11.35 \\

SVM (polynomial) & 80.61 & 72.71 & 78.61 & 45.62 & 62.39 & 61.95 & 17.57 \\

RF (10) & 78.40 & 68.87 & 55.75 & 51.47 & 46.80 & 48.01 & 32.35 \\

RF (50) & 89.27 & 34.38 & 77.75 & 72.31 & 68.65 & 70.48 & 46.96 \\

RF (100) & 90.65 & 86.88 & 84.29 & 78.87 & 76.23 & 77.32 & 54.16 \\

3-NN & 77.39 & 82.63 & 75.41 & 68.93 & 65.59 & 64.23 & 39.13 \\

5-NN & 75.28 & 80.99 & 73.18 & 66.49 & 63.33 & 62.27 & 38.46 \\

7-NN & 72.93 & 79.11 & 70.58 & 64.44 & 60.74 & 59.60 & 36.74 \\
    \hline
    \multicolumn{8}{c}{\textbf{EuroSAT 5-way/5-shot}} \\
    \textbf{Learner} & \textbf{Ensemble} & \textbf{Output} & \textbf{Block 8} & \textbf{Block 7} & \textbf{Block 6} & \textbf{Block 5} & \textbf{Block 4} \\
    \hline
Hebbian & 83.34 & 80.87 & 82.15 & 80.49 & 76.51 & 74.89 & 59.38 \\

SVM (linear) & 80.54 & 78.73 & 78.20 & 71.78 & 67.05 & 68.35 & 46.37 \\

SVM (RBF) & 57.82 & 63.79 & 39.44 & 30.75 & 31.42 & 33.31 & 26.28 \\

SVM (polynomial) & 60.99 & 53.68 & 39.97 & 27.97 & 52.90 & 55.60 & 33.28 \\

RF (10) & 70.29 & 64.32 & 51.94 & 49.35 & 47.52 & 47.35 & 36.34 \\

RF (50) & 80.07 & 74.59 & 67.37 & 65.46 & 61.93 & 63.03 & 45.61 \\

RF (100) & 81.60 & 76.52 & 71.98 & 71.77 & 67.61 & 68.90 & 50.85 \\

3-NN & 39.99 & 67.25 & 42.93 & 33.69 & 40.42 & 38.93 & 21.24 \\

5-NN & 37.27 & 65.19 & 38.36 & 31.33 & 39.36 & 37.67 & 20.95 \\

7-NN & 72.93 & 61.95 & 34.06 & 28.98 & 38.27 & 36.61 & 21.11 \\
    \hline
    \multicolumn{8}{c}{\textbf{ISIC 5-way/5-shot}} \\
    \textbf{Learner} & \textbf{Ensemble} & \textbf{Output} & \textbf{Block 8} & \textbf{Block 7} & \textbf{Block 6} & \textbf{Block 5} & \textbf{Block 4} \\
    \hline
Hebbian  & 46.29 & 42.66 & 45.03 & 42.51 & 42.35 & 42.69 & 37.49 \\

SVM (linear) & 30.85 & 35.49 & 33.92 & 26.82 & 23.66 & 23.76 & 15.40 \\

SVM (RBF) & 13.83 & 16.65 & 15.33 & 12.30 & 12.19 & 11.94 & 13.83 \\

SVM (polynomial) & 18.17 & 20.81 & 21.65 & 13.66 & 18.30 & 18.90 & 12.90 \\

RF (10) & 37.06 & 33.76 & 30.06 & 27.21 & 26.18 & 25.86 & 23.31 \\

RF (50) & 42.12 & 39.47 & 37.75 & 33.68 & 31.54 & 32.15 & 25.63 \\

RF (100) & 44.04 & 41.20 & 40.00 & 36.68 & 34.80 & 35.62 & 28.51 \\

3-NN & 35.24 & 34.33 & 33.11 & 29.34 & 30.05 & 30.38 & 23.23 \\

5-NN & 34.49 & 34.32 & 32.62 & 29.37 & 30.12 & 30.11 & 23.14 \\

7-NN & 32.86 & 33.24 & 31.70 & 28.35 & 29.22 & 29.73 & 22.35 \\
    \hline
    \multicolumn{8}{c}{\textbf{ChestX 5-way/5-shot}} \\
    \textbf{Learner} & \textbf{Ensemble} & \textbf{Output} & \textbf{Block 8} & \textbf{Block 7} & \textbf{Block 6} & \textbf{Block 5} & \textbf{Block 4} \\
    \hline
Hebbian & 26.11 & 24.79 & 25.70 & 25.30 & 24.93 & 24.83 & 23.61 \\

SVM (linear) & 18.39 & 20.54 & 19.03 & 18.28 & 18.11 & 18.24 & 18.36 \\

SVM (RBF) & 16.27 & 16.73 & 15.97 & 17.07 & 17.61 & 17.71 & 18.09 \\

SVM (polynomial) & 16.83 & 17.30 & 18.39 & 17.23 & 17.69 & 17.62 & 18.21 \\

RF (10) & 23.52 & 22.05 & 22.39 & 21.64 & 21.51 & 21.69 & 21.18 \\

RF (50) & 25.47 & 23.57 & 23.56 & 23.62 & 23.22 & 23.02 & 21.74 \\

RF (100) & 26.24 & 24.32 & 25.10 & 24.24 & 24.12 & 23.98 & 22.51 \\

3-NN & 24.47 & 22.89 & 24.02 & 23.57 & 23.00 & 22.66 & 21.72 \\

5-NN & 24.15 & 23.32 & 24.11 & 23.65 & 22.94 & 23.00 & 21.53 \\

7-NN & 24.57 & 23.25 & 24.48 & 23.02 & 23.21 & 23.13 & 21.77 \\
    \hline
\end{tabular}}
\caption{Results of our few-shot learning method CHEF on four proposed cross-domain few-shot challenges CropDiseases, EuroSAT, ISIC, and ChestX using the ResNet-18 backbone pre-trained on Imagenet. We compare Hebbian learning to support vector machine (SVM) with linear, polynomial and RBF kernel, respectively alongside random forest (RF) with 10, 50, and 100 trees as well as $k$ nearest neighbor (NN) with $k=3$, $k=5$, and $k=7$, respectively. For each learner, we compare ensemble results to the results obtained when using only the features of the abstraction layer at the end of the blocks 4, 5, 6, 7, and 8 as well as the output layer. All learners were used on top of a ResNet-18 backbone network pre-trained on Imagenet. The average 5-way few-shot classification accuracies ($\%$, top-1) over 800 episodes are reported.}
\label{tab:ablation2}
\end{table*}

\subsection{Computation time}
Although the presented algorithm does not differ from most other few-shot or meta-learning methods in terms of computational complexity, which is $O(n)$, it is still among the fastest algorithms in this realm. The main reason for this is that after pre-training our algorithm does not perform episodic training but only evaluation. That is, no information is carried from one episode to another in the target domain. Therefore, CHEF only needs to activate the backbone architecture and then it uses the features of certain layers as inputs for the Hebbian ensemble learner. In particular, it does not apply backpropagation after pre-training, i.e.\ there is no need for differentiating the learner.
The execution times on a Tesla P100 16GB for episodes with 5/20/50 shots are 1.258/1.909/3.504 seconds on average over 10 runs, respectively. On the same infrastructure, but utilizing 2 cards in parallel, pre-training one epoch on \textit{mini}Imagenet takes 48.54 and 82.73 seconds for the Conv64 and ResNet-12 backbone, respectively, where one epoch contains 159 update steps. Pre-training one epoch on \textit{tiered}Imagenet takes 113.56 and 973.33 seconds for the Conv64 and ResNet-12 backbone, respectively, where one epoch contains 1,902 update steps. For both datasets, we performed 500 epochs of pre-training. 

\subsection{Experimental setup}
In the following, we give further details on our experimental setups. 

\subsubsection{Cross-domain few-shot learning}
We utilize a ResNet-10 backbone architecture as proposed in~\citet{guo19}. 
The residual blocks have 64, 128, 256, 512, 4000, and 1000 units, where the 
latter two are fully connected ReLU layers. 
We use a learning rate of 0.1, 
momentum term of 0.9, L2 weight decay term of $10^{-4}$, batch size of 256, 
dropout rate of 0.5 during pre-training. 
These values were tuned on the horizontal validation set of \textit{mini}Imagenet. 
For few-shot learning, we choose a Hebbian learning rate of $\alpha=0.01$ 
and run the Hebb rule for $I=400$ steps. 
These values were tuned on the vertical validation set of \textit{mini}Imagenet. 

\begin{figure}[!ht]
\begin{center}
\includegraphics[width=.5\textwidth]{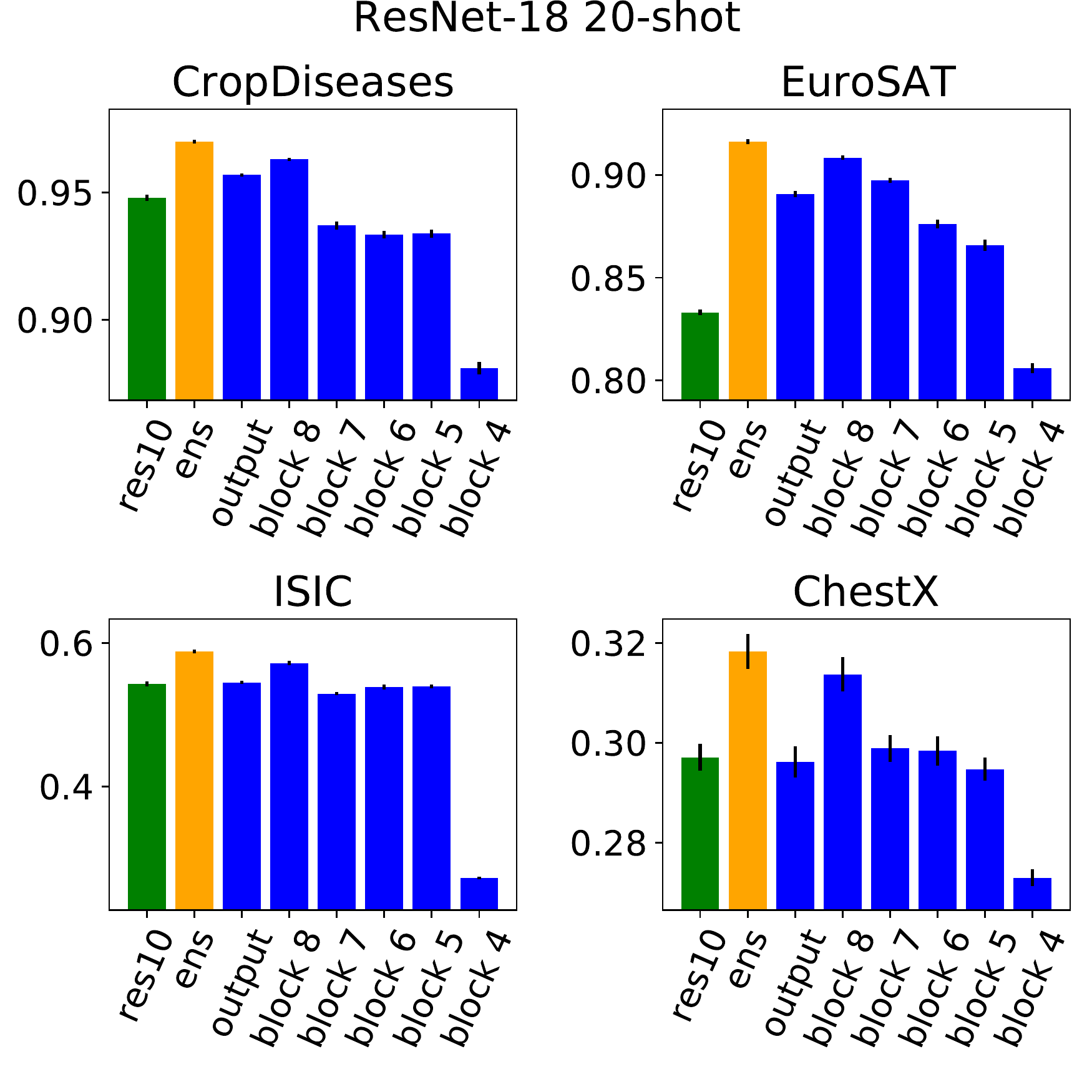}
\end{center}
\caption{20-shot top-1 accuracies (along with $95\%$ confidence intervals) 
of different residual blocks and the output layer of an 
Imagenet-pretrained ResNet-18 and the ensemble result (orange, ``ens'') on 
the four different datasets 
of the cross-domain few-shot learning benchmark. 
For comparison, also the ResNet-10 ensemble results (green) are included.}
\label{fig:res18_20shot}
\end{figure}

Figures \ref{fig:res18} and \ref{fig:res18_20shot} show the
performance of the pre-trained PyTorch ResNet-18 network,
where the pre-training is on the entire Imagenet dataset.
Additionally, the individual performances of the ResNet-18
layers are depicted.
The \textit{mini}Imagenet pre-trained ResNet-10 is shown for comparison.
%performance of the ResNet-10, which was pre-trained on \textit{mini}Imagenet
%in comparison to a ResNet-18, which was pre-trained on the entire Imagenet dataset 
%on the four cross-domain 5-shot, 20-shot, and 50-shot tasks, respectively. 
%Additionally, the individual performances of the ResNet-18 layers is depicted. 
The plots show the general tendency that the ensemble performance on domains 
which are farther away from the training domain relies more heavily on features 
in lower layers, i.e.\ features with less specificity to the original domain.

\subsubsection{miniImagenet and tieredImagenet}
\paragraph{Backbone pre-training.}
For the \textit{mini}Imagenet and \textit{tiered}Imagenet experiments,
we utilize Conv-4 and ResNet-12 architectures as backbone networks.
The Conv-4 network is described in detail by \citet{vinyals16}.
It is a stack of 4 modules,
each of  which consists of a $3 \times 3$ convolutional layer with 64 units, 
a batch normalization layer \citep{ioffe15},
a ReLU activation and $2 \times 2$ max-pooling layer. On top, 
we place two fully connected ReLU layers with 400 and 100 units, respectively. 
The ResNet-12 is described in \citet{lee19}.
We configure the backbone as 4 residual blocks 
with 64, 160, 320, 640, 4000, and 1000 units, respectively,
where the latter two are ReLU-activated fully connected layers. 
The residual blocks contain a max-pooling and a batch-norm layer and are regularized by DropBlock \citep{ghiasi18}
with block sizes of $1 \times 1$ for the first two blocks and $5 \times 5$ for the latter two blocks. 

We pre-train these backbone models for 500 epochs with three different learning rates $0.1$, $0.01$, and $0.001$.
For this we use the PyTorch SGD module for stochastic gradient descent
with a momentum term of $0.9$, 
an L2 weight decay factor of $10^{-4}$, 
a mini-batchsize of $256$, 
and a dropout probability of $0.1$.
This pre-training is performed on the horizontal training set of the \textit{mini}Imagenet and the \textit{tiered}Imagenet dataset,
resulting in 3 trained models per dataset.
We apply early stopping by selecting the model with the lowest loss on the horizontal validation set,
while evaluating the model performance after each epoch.

\paragraph{Few-shot learning.}
For few-shot learning, we perform a grid search to determine the best hyper-parameter setting for each of the datasets and each of the 1-shot and 5-shot settings,
using the loss on the vertical validation set.
We treat the 3 backbone models that were pre-trained with different learning rates, 
as described in the previous paragraph,
as hyper-parameters.
The hyper-parameters used for this grid-search are listed in
Tab.~\ref{tab:mini_tiered_search_space}.

After determining the best hyper-parameter setting following this procedure,
we perform 1-shot and 5-shot learning on the vertical test sets of \textit{mini}Imagenet 
and \textit{tiered}Imagenet using $10$ different random seeds, respectively.
The results are listed in Tab.~\ref{tab:main}.

\begin{table*}[!p]
\centering
% \scalebox{0.75}{
\begin{tabular}{lc}
\hline 
\textbf{parameter} & \textbf{values} \\
\hline 
learning rate of pre-trained model & $\{0.1, 0.01, 0.001\}$ \\
dropout probability & $0.5$ \\
Hebbian learning rate & $\{0.1, 0.01, 0.001\}$ \\
number of Hebb rule steps & $\{1, 2, 5, 7, 10, 25, 50, 75, 100, 250, 500, 750\}$ \\
\hline
\end{tabular}%
% }
\caption{Hyper-parameter search space for 1-shot and 5-shot learning on \textit{mini}Imagenet and \textit{tiered}Imagenet using Conv-4 and ResNet-12 backbone models.
Best hyper-parameters were evaluated using a grid-search and the loss on the vertical validation set of \textit{mini}Imagenet or \textit{tiered}Imagenet.}
\label{tab:mini_tiered_search_space}
% \vspace{-0.16in}
\end{table*}

\begin{figure}[!ht]
\begin{center}
\includegraphics[width=0.66\textwidth]{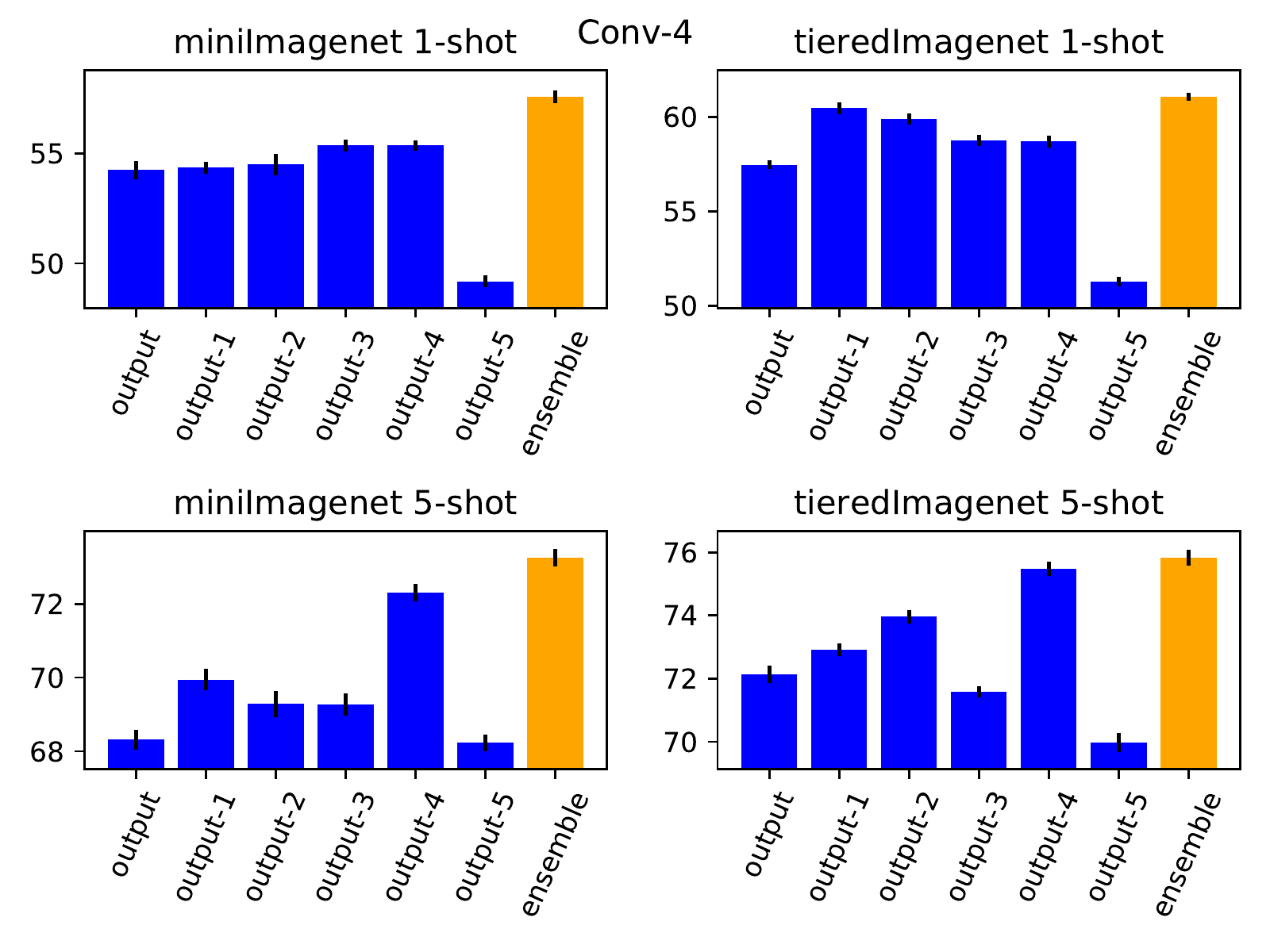}
\end{center}
\caption{Ablation study of the Conv-4 architecture on the \textit{mini}Imagenet
and \textit{tiered}Imagenet datasets for 1-shot and 5-shot. The plots show the 
individual performances of Hebbian learners acting on single layers and their 
ensemble performance along with 95\% confidence intervals. The labels on the 
$x$-axis indicate how far the respective 
layer is from the output layer.}
\label{fig:conv64}
\end{figure}

\begin{figure}[!ht]
\begin{center}
\includegraphics[width=0.66\textwidth]{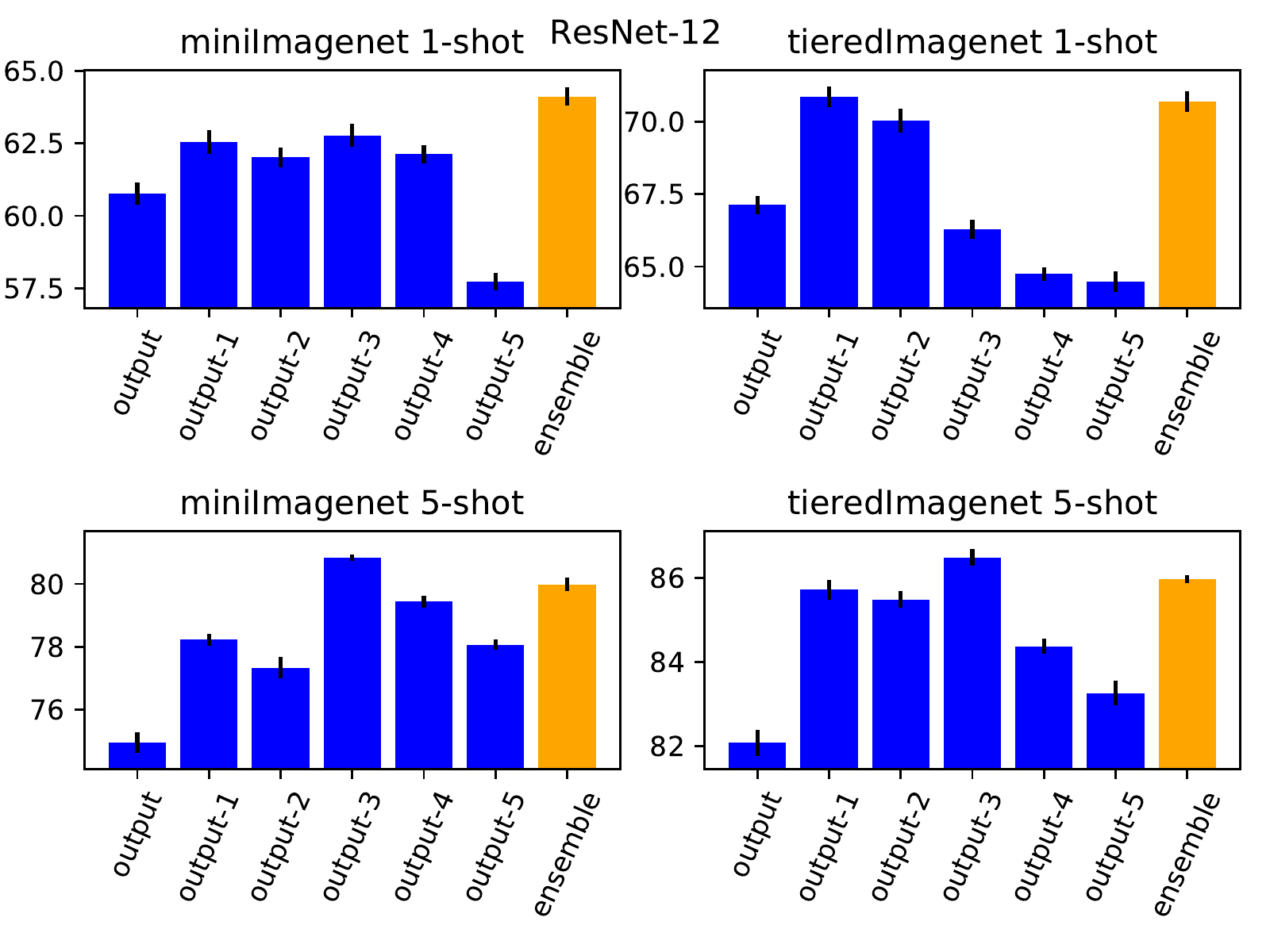}
\end{center}
\caption{Ablation study of the ResNet-12 architecture on the \textit{mini}Imagenet
and \textit{tiered}Imagenet datasets for 1-shot and 5-shot. The plots show the 
individual performances of Hebbian learners acting on single layers and their 
ensemble performance along with 95\% confidence intervals. The labels on the 
$x$-axis indicate how far the respective 
layer is from the output layer.}
\label{fig:res12}
\end{figure}

\paragraph{Ensemble learning and performance of individual layers.}
To evaluate the performance of the Hebbian learning using only individual layers versus using the ensemble of layers,
we additionally perform the few-shot learning on the vertical test sets using only individual layers as input to the Hebbian learning.
As shown in Figures~\ref{fig:conv64} and~\ref{fig:res12},
the performance using only the individual layers varies strongly throughout 1-shot and 5-shot learning and the \textit{mini}Imagenet and \textit{tiered}Imagenet dataset.
This indicates that the usefulness of the representations provided by the individual layers strongly depends on the data and task setting.
In contrast to this,
the ensemble of layers reliably achieves either best or second best performance throughout all settings.

%\clearpage

\subsubsection{Example Application: Drug Discovery}

%\paragraph{Details on the evaluation procedure}
%The evaluation is done considering
%a 50-shot 2-way few-shot learning task,
%For balanced data, the training set 
%would comprise 50 samples for each of the two classes (toxic and non-toxic).
%To account for imbalance, this number is adapted, 
%such that the imbalance of the assays is reflected: For the class with fewer
%samples the number of samples is reduced, while the number of samples for the class
%with an overall higher number of samples is increased.
%For all experiments, we use 20 samples from each class in the test set.
%We repeat the process of drawing samples for training and test sets for 100 times. 
%We randomly flip the class labels of the created training set in order 
%to avoid biases that could be 
%introduced by molecules that are active or inactive in most of the assays.

\paragraph{Details on pre-training on the ChEMBL20 database}
For training a fully-connected deep neural network (FCN)
on the ChEMBL20 database,
the number of 892,480 features  is reduced
by a sparseness criterion on the molecules to 1,866 features.
The FCN is trained on 1.1 million molecules for 1,000 epochs 
minimizing binary cross-entropy and masking out missing values by using an objective,
as described in \citep{mayr2018large}.

\paragraph{Details on the compared methods}
We use ECFP6 features for a raw molecule representation.
Note that the number of possible distinct ECFP6 features is not predefined, since a new 
molecule may be structurally different from all previously seen ones, and it might therefore
consist of new unseen ECFP6 features.
For SVMs, a MinMax kernel \citep{mayr2016deeptox} is used,
which operates directly on counts of ECFP6 features
and used LIBSVM \citep{chang2011libsvm} 
as provided by \texttt{scikit-learn} \citep{scikit-learn}.
For RFs, the implementation of \texttt{scikit-learn} 
with 1000 trees and kept default values for
the other hyperparameters is used.

\paragraph{Detailed results on the Tox21 dataset in a few-shot setup}
Table \ref{tab:tox21detail} lists detailed results and
$p$-values of all twelve few-shot tasks of the Tox21 Data Challenge.
For calculating $p$-values, a paired Wilcoxon test is used.

\begin{table*}[!p]
\centering
\scalebox{0.80}{
\begin{tabular}{l c c c c c c}
    \hline
    %\hline
\textbf{Dataset} &
\textbf{CHEF} & 
\textbf{SVM} & 
\textbf{RF} & 
\textbf{p}-\textbf{value SVM} & 
\textbf{p}-\textbf{value RF} \\
\hline

NR.AhR        &  \textbf{0.86 $\pm$ 0.07} &  0.79 $\pm$ 0.07 &  0.75 $\pm$ 0.07 &         2.90e-12 &        1.19e-17 \\
NR.AR         &  \textbf{0.79 $\pm$ 0.09} &  0.60 $\pm$ 0.11 &  0.61 $\pm$ 0.11 &         1.20e-17 &        5.25e-18 \\
NR.AR.LBD     &  \textbf{0.84 $\pm$ 0.05} &  0.47 $\pm$ 0.11 &  0.52 $\pm$ 0.10 &         1.94e-18 &        1.95e-18 \\
NR.Aromatase  &  \textbf{0.74 $\pm$ 0.08} &  0.68 $\pm$ 0.09 &  0.64 $\pm$ 0.09 &         3.77e-09 &        1.12e-13 \\
NR.ER         &  \textbf{0.73 $\pm$ 0.08} &  0.70 $\pm$ 0.08 &  0.65 $\pm$ 0.09 &         1.39e-03 &        4.25e-11 \\
NR.ER.LBD     &  \textbf{0.71 $\pm$ 0.08} &  0.68 $\pm$ 0.09 &  0.65 $\pm$ 0.10 &         1.96e-03 &        2.40e-06 \\
NR.PPAR.gamma &  \textbf{0.66 $\pm$ 0.07} &  0.61 $\pm$ 0.10 &  0.60 $\pm$ 0.11 &         8.04e-06 &        3.05e-06 \\
SR.ARE        &  \textbf{0.76 $\pm$ 0.08} &  0.66 $\pm$ 0.08 &  0.61 $\pm$ 0.09 &         2.43e-14 &        2.51e-17 \\
SR.ATAD5      &  \textbf{0.68 $\pm$ 0.07} &  0.62 $\pm$ 0.10 &  0.61 $\pm$ 0.10 &         2.23e-07 &        7.65e-10 \\
SR.HSE        &  \textbf{0.74 $\pm$ 0.06} &  0.62 $\pm$ 0.10 &  0.60 $\pm$ 0.10 &         3.42e-16 &        1.40e-16 \\
SR.MMP        &  \textbf{0.89 $\pm$ 0.05} &  0.81 $\pm$ 0.08 &  0.79 $\pm$ 0.09 &         7.36e-15 &        4.41e-16 \\
SR.p53        &  \textbf{0.77 $\pm$ 0.08} &  0.67 $\pm$ 0.10 &  0.63 $\pm$ 0.10 &         3.50e-13 &        8.86e-17 \\

\hline
\end{tabular}

}
\caption{
ROC-AUC performances for the twelve individual few-shot tasks (rows) of the Tox21 Data Challenge.
CHEF is compared to conventional methods (SVM, RF).
Averages and standard deviations are computed across 100 differently sampled training and test sets.
The last two columns show the results of paired Wilcoxon tests with the null hypotheses given that SVM and RF perform better, respectively.
}
\label{tab:tox21detail}
\end{table*}

\end{document}

%% file: math_commands.tex
\usepackage{amsmath,amsfonts,bm}

\def\eqref#1{equation~\ref{#1}}
\def\1{\bm{1}}

\DeclareMathAlphabet{\mathsfit}{\encodingdefault}{\sfdefault}{m}{sl}
\SetMathAlphabet{\mathsfit}{bold}{\encodingdefault}{\sfdefault}{bx}{n}